\documentclass[11pt]{article}

\usepackage[final]{acl}

\usepackage{times}
\usepackage{latexsym}

\usepackage[T1]{fontenc}

\usepackage[utf8]{inputenc}
\usepackage{tabularx}

\newcolumntype{Y}{>{\centering\arraybackslash}X}
\usepackage{microtype}

\usepackage{inconsolata}

\usepackage{graphicx}

\newcommand{\agent}[1]{\path{#1}}
\newcommand{\dagcell}[1]{\path{#1}}
\newcommand{\mdedge}{\ensuremath{\mathrm{M{\to}D}}}
\newcommand{\dmedge}{\ensuremath{\mathrm{D{\to}M}}}

\usepackage{amsmath,amssymb}
\usepackage{booktabs}
\usepackage{multirow}
\usepackage{array}
\usepackage{listings}
\usepackage{enumitem}
\title{DisasterBench: Benchmarking LLM Planning under\\ Typed Tool Interface Constraints
}

\author{
\textbf{Zhitong Chen}\textsuperscript{*1}\quad
\textbf{Kai Yin}\textsuperscript{*1}\quad
\textbf{Weifeng Zhang}\textsuperscript{1}\quad
\textbf{Zhiyuan Wang}\textsuperscript{1}\quad
\textbf{Xiangjue Dong}\textsuperscript{1}\quad
\textbf{Chengkai Liu}\textsuperscript{1}\\
\textbf{Zhewei Liu}\textsuperscript{2}\quad
\textbf{Yiming Xiao}\textsuperscript{1}\quad
\textbf{Ali Mostafavi}\textsuperscript{1}\quad
\textbf{James Caverlee}\textsuperscript{1}\\
\textsuperscript{1}Texas A\&M University \quad
\textsuperscript{2}University of Toronto\\
\texttt{\{zhitong.chen18,kai.yin,weifengzhang,zhiyuan,xj.dong\}@tamu.edu}\\
\texttt{lzwgre@gmail.com}\\
\texttt{\{yxiao,mostafavi,caverlee\}@tamu.edu}
}

\begin{document}
\maketitle
\begingroup
\renewcommand\thefootnote{\fnsymbol{footnote}}
\footnotetext[1]{Equal contribution.}
\footnotetext[2]{Corresponding authors.}
\endgroup
\maketitle
\begin{abstract}
Disasters cause severe societal impacts, demanding rapid coordination of heterogeneous AI tools, from satellite analysis to flood prediction and damage assessment, into coherent multi-step workflows. As LLMs increasingly serve as orchestrators of such pipelines, effective coordination requires more than selecting semantically plausible tools: LLMs must generate executable workflows with correct parameter binding and dependency propagation. We introduce \textbf{DisasterBench}, a benchmark for evaluating structured
multi-agent planning over semantically similar but operationally distinct
disaster-response tools. To enable step-level failure attribution, we
further propose \textbf{First-Point-of-Failure (FPoF)}, which localizes
the earliest root cause in a predicted workflow, separating primary
errors from downstream cascading effects. Our evaluation reveals three
findings: planning method effectiveness depends strongly on model
capacity; tool mismatch and parameter-binding errors dominate first
failures, revealing semantic grounding and execution consistency as
distinct bottlenecks; and verbose intermediate reasoning can create
instruction clash with structured output requirements,
disrupting plan generation. Together, these findings highlight a
fundamental gap between semantic reasoning and execution-grounded
coordination, underscoring the need for planning frameworks that jointly
model semantic intent, execution constraints, and workflow consistency.
All code, data, and evaluation resources are available at the
\href{https://github.com/TamuChen18/DisasterBench_Open}{project page}.
\end{abstract}

\begin{figure*}[t]
    \centering
    \includegraphics[width=\textwidth]{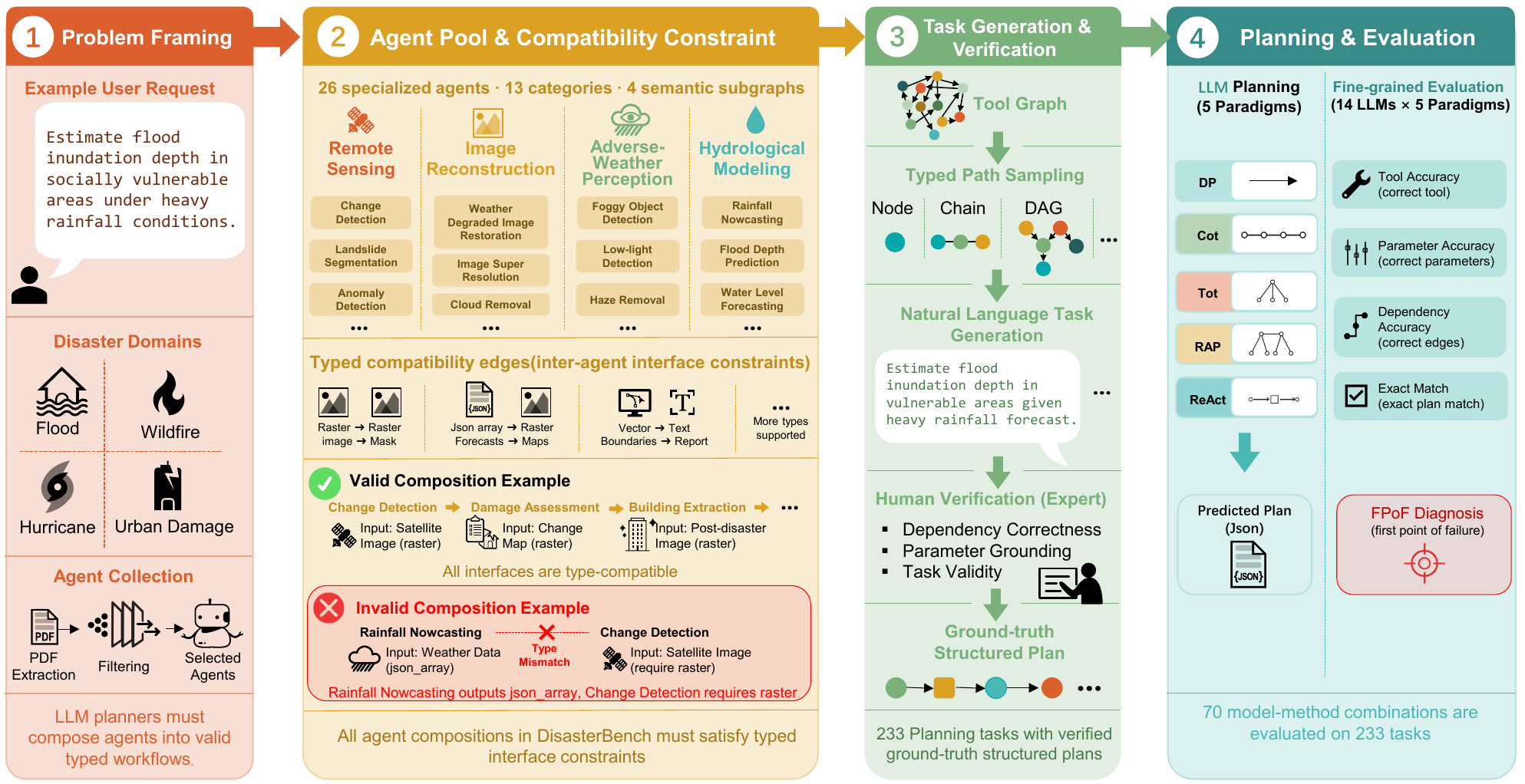}
    \caption{Overview of the DisasterBench benchmark construction and 
    evaluation pipeline, comprising four stages: tool pool construction 
    with typed inter-agent interface constraints, workflow execution 
    constraint definition, natural-language task generation with expert 
    verification, and multi-paradigm LLM evaluation with FPoF diagnosis.}
    \label{fig:disasterbench_pipeline}
\end{figure*}

\section{Introduction}

Modern disaster response increasingly relies on coordinating
heterogeneous AI capabilities into coherent multi-stage workflows.
Emergency management agencies, first responders, and domain analysts must rapidly orchestrate specialized tools for tasks such as satellite image analysis, precipitation nowcasting, flood modeling, and damage assessment. These workflows are operationally constrained: outputs from one stage often serve as structured inputs to downstream stages, making workflow correctness dependent on both semantic grounding and execution consistency.
Recent advances in LLM-powered multi-agent systems
~\citep{tran2025multiagent} have created new opportunities for
automating such workflow orchestration. However, effective deployment requires reliable \textit{workflow grounding}, where models must not only identify relevant tools but also generate executable workflows with correct parameter bindings and inter-step dependencies.

Despite recent progress in multi-agent benchmarking
~\citep{liu2024agentbench, shen2024taskbench} and tool-augmented LLMs
~\citep{qin2024toolllm, patil2024gorilla}, existing benchmarks provide
limited evaluation of workflow grounding in disaster-response settings.
This limitation is especially critical because disaster-response tools
are often semantically similar but operationally distinct. For example,
multiple flood-analysis tools may appear equally relevant for a task
while requiring different inputs, spatial assumptions, or downstream
compatibility. Selecting an operationally incompatible tool may
therefore invalidate downstream workflow stages even when the
high-level intent appears reasonable.

To address these challenges, we introduce \textbf{DisasterBench}, a
benchmark for evaluating workflow grounding in disaster-response
multi-agent systems. DisasterBench is built around 26 openly available
disaster-response tools spanning satellite image analysis,
adverse-weather perception, hydrological modeling, and post-disaster
mobility prediction. From expert-designed disaster-response workflows,
we derive 233 planning tasks ranging from single-step perception to
multi-stage branching workflows. Models are provided only with tool
specifications and interface schemas, without access to the underlying
workflow structure, requiring them to infer executable plans directly
from natural-language task descriptions.
To enable precise failure attribution beyond aggregate metrics, we
further propose \textbf{First-Point-of-Failure (FPoF)}, a diagnostic
framework that localizes the earliest root cause in a generated
workflow, separating primary failures from cascaded downstream errors.

Our evaluation of 14 LLMs across five planning paradigms yields three
primary findings. First, plan depth is a universal bottleneck:
performance degrades sharply beyond two steps across all models and methods, suggesting that multi-step coordination imposes compounding difficulty that increased model capacity alone does not resolve.
Second, tool mismatch is the dominant workflow-grounding failure,
revealing that distinguishing among semantically similar but
operationally distinct tools remains a major challenge. Third, verbose
intermediate reasoning can exhibit instruction clash with structured
output requirements, disrupting workflow generation rather than
improving it.
Our contributions are as follows:

\textbf{DisasterBench.}
We introduce DisasterBench, a benchmark for evaluating workflow
grounding in disaster-response multi-agent systems.

\textbf{First-Point-of-Failure (FPoF).}
We propose a step-level diagnostic framework for localizing root
causes in predicted workflows.

\textbf{Empirical Analysis.}
We provide a systematic empirical analysis of structured multi-agent
planning failures, identifying bottlenecks including depth-driven
collapse, workflow-grounding failures, and instruction clash.

\begin{table*}[t]
\centering
\small
\setlength{\tabcolsep}{5pt}
\renewcommand{\arraystretch}{1.02}
\begin{tabularx}{\textwidth}{|>{\centering\arraybackslash}m{0.20\textwidth}|Y|>{\centering\arraybackslash}m{0.04\textwidth}|}
\hline
\textbf{Category} & \textbf{Tool(s)} & \textbf{\#} \\
\hline
Image Gen./Restoration & {\small\ttfamily Temporal\_High\_Resolution\_Image\_Generation, High-Resolution\_Image\_Reconstructor, Metadata\_and\_Text\_Prompt\_Image\_Generation, Weather\_Degraded\_Image\_Restoration} & 4 \\
\hline
Classification & {\small\ttfamily Temporal\_Image\_Sequence\_Classifier, RGB\_GeoImage\_Classifier, Multi\_Spectral\_Classifier} & 3 \\
\hline
Anomaly Detection & {\small\ttfamily Anomaly\_Detection\_Forest, Landslide\_Segmentation, Urban\_Anomaly\_Detection} & 3 \\
\hline
Change Detection & {\small\ttfamily Change\_Mapping\_and\_Detection} & 1 \\
\hline
Object Det./Seg. & {\small\ttfamily Foggy\_Scenario\_Object\_Detection, Low-Light\_Object\_Detection, Geospatial\_Object\_Segmentation} & 3 \\
\hline
Crowd Counting & {\small\ttfamily Crowd\_Counting\_in\_Adverse\_Weather} & 1 \\
\hline
Building Damage & {\small\ttfamily Building\_damage\_assessment} & 1 \\
\hline
Flood \& Precipitation & {\small\ttfamily Precipitation\_Nowcasting, Flood\_depth\_prediction, depth\_speed\_model} & 3 \\
\hline
Text \& Event Proc. & {\small\ttfamily Toponym\_Detection, Event\_detection} & 2 \\
\hline
Video Analysis & {\small\ttfamily Video\_anomaly\_detection} & 1 \\
\hline
Mobility Prediction & {\small\ttfamily Post\_Disaster\_Mobility\_Recovery, Multimodal\_mobility\_prediction\_under\_events} & 2 \\
\hline
Multimodal Reasoning & {\small\ttfamily GeoChat} & 1 \\
\hline
Data Conversion & {\small\ttfamily precipitation\_data\_convert\_tool} & 1 \\
\hline
\textbf{Total} & & \textbf{26} \\
\hline
\end{tabularx}
\caption{Summary of the 26 specialized agents in DisasterBench,
grouped into 13 functional categories spanning perception,
prediction, generation, and reasoning tasks in
disaster-response workflows.}
\label{tab:agent_pool}
\end{table*}

\section{Related Work}

\subsection{Tool-Augmented and Multi-Step Planning Benchmarks}

Prior benchmarks such as ToolBench~\citep{qin2024toolllm},
APIBench~\citep{patil2024gorilla}, AgentBench~\citep{liu2024agentbench},
and TaskBench~\citep{shen2024taskbench} evaluate LLMs on tool selection,
invocation, and multi-step workflow orchestration, but treat agents as
freely composable without evaluating whether resulting workflows remain
executable under downstream compatibility and execution constraints.
DisasterBench instead evaluates executable workflow grounding over
semantically similar but operationally distinct tools, and provides
step-level failure attribution through FPoF to diagnose why semantically
plausible workflows fail to remain executable.

\subsection{Structured Generation and Executable Planning}

Structured generation and semantic parsing study how models map
natural-language intent into executable representations~\citep{yin2017syntactic,
scholak2021picard, shin2021constrained}, while reasoning paradigms such
as CoT~\citep{wei2022chain}, ToT~\citep{yao2023tree},
ReAct~\citep{yao2022react}, and RAP~\citep{hao2023reasoning} improve
multi-step planning through explicit intermediate reasoning. DisasterBench
is complementary to both lines of work: it evaluates executable workflow
grounding beyond syntactic validity alone, and reveals that verbose
reasoning traces can interfere with structured executable generation.
We term this phenomenon \emph{instruction clash}.

\subsection{AI for Disaster Management}

Prior work has advanced individual disaster-response capabilities
including remote-sensing-based disaster analysis~\citep{algiriyage2022multi}, flood mapping and forecasting~\citep{bentivoglio2022deep}, and damage assessment~\citep{gupta2019xbd},
with recent benchmarks evaluating disaster-domain QA~\citep{chen2026disastqa}
and text-to-SQL~\citep{liu2025floodsql}. DisasterBench shifts focus to
multi-agent workflow grounding, studying whether LLMs can coordinate
semantically similar but operationally distinct disaster-response tools
into execution-consistent workflows and diagnosing why fail.

\section{Benchmark Construction}
\label{sec:benchmark_construction}

Disaster-response workflows present a distinctive multi-agent coordination
challenge: tools are often functionally overlapping but operationally
distinct, and valid workflows must maintain consistent parameter bindings
and inter-step dependencies across multiple stages. As a result, workflow
correctness depends not only on selecting plausible tools, but also on
generating execution-consistent plans under downstream execution
constraints. DisasterBench is designed to systematically evaluate this
workflow grounding problem. Unlike existing benchmarks that primarily
evaluate tool invocation or workflow decomposition, DisasterBench requires
models to infer executable workflows over structured disaster-response
pipelines without access to the workflow structure.
Figure~\ref{fig:disasterbench_pipeline} provides an overview of the
benchmark construction pipeline; the following subsections detail each
component.

\subsection{Tool pool}
\label{sec:agent_pool}

We construct a pool of 26 specialized tools spanning 13 functional
categories covering core perception, prediction, and reasoning tasks in
disaster-response workflows (Table~\ref{tab:agent_pool}). These agents
span heterogeneous modalities and task settings, from satellite image
analysis and precipitation nowcasting to adverse-weather perception,
damage assessment, and post-disaster mobility modeling, reflecting the
diverse capabilities required in disaster-response workflows.

The pool is designed to probe fine-grained workflow grounding rather
than simple tool retrieval. Many tools are \textit{functionally
overlapping but operationally distinct}: forest anomaly detection and
urban anomaly detection both identify abnormal visual patterns but apply
to different disaster contexts; foggy-object detection and
low-light-object detection both target degraded visual conditions but
require different operational assumptions. Such overlaps require models
to perform precise semantic disambiguation while preserving valid
downstream dependencies. The difficulty of DisasterBench arises not
from the number of tools, but from the high semantic similarity among
operationally distinct candidates, as further supported by the
tool-grounding failure patterns in Section~\ref{sec:failure_modes}
and quantified through description-level semantic similarity analysis
(Appendix~\ref{app:agent_overlap}).

Each tool is described by its functional role and input/output schema,
defining the structured fields used in ground-truth workflows and model
predictions. Tools are sourced from publicly available models reported
in top-tier venues, selected through semi-automated filtering and manual
verification for domain relevance, functional diversity, and
implementation availability
(Appendix~\ref{app:filter_prompt}).
 
\begin{figure}[t]
    \centering
    \IfFileExists{depth_accuracy.pdf}{%
        \includegraphics[width=\columnwidth]{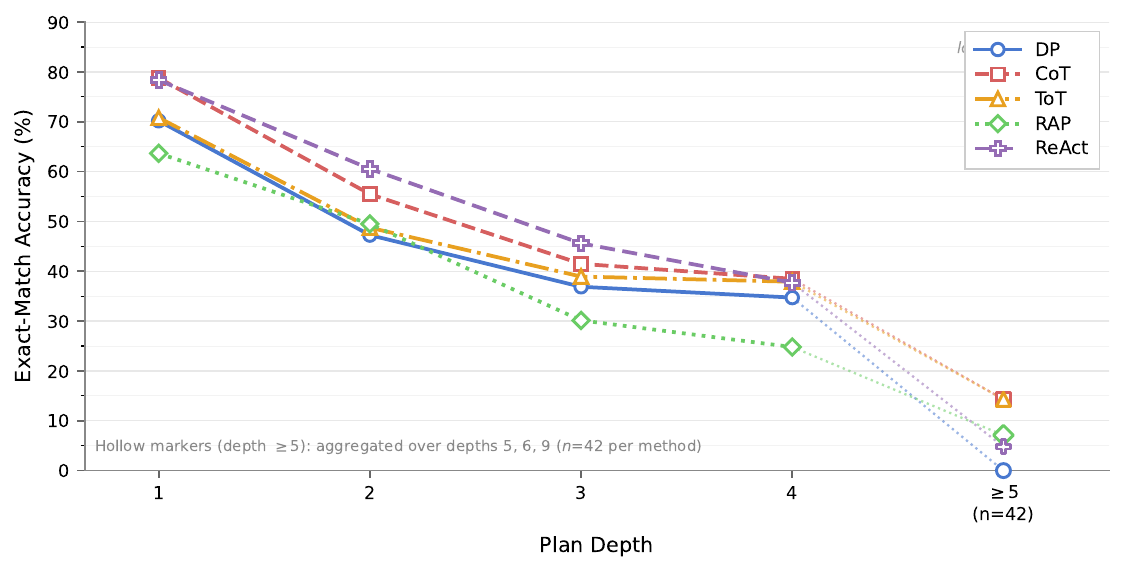}%
    }{%
        \fbox{\parbox[c][0.28\textheight][c]{0.95\columnwidth}{\centering Depth accuracy figure placeholder\\(add \texttt{figures/depth\_accuracy.pdf})}}%
    }
    \caption{Exact-match accuracy (\%) as a function of ground-truth plan depth,
             micro-averaged over all models for each planning method. All methods exhibit a sharp performance cliff as workflow depth increases, confirming that execution-consistency requirements compound with plan depth in DisasterBench.}
    \label{fig:depth_accuracy}
\end{figure}

\subsection{Workflow Execution Constraints}
\label{sec:workflow_dependency}

We define an expert-curated set of workflow execution constraints over
the 26 tools, specifying which tool outputs are admissible inputs to
downstream stages. The resulting structure contains 81 directed
dependencies organized into four coherent disaster-response workflow
families; full details are provided in
Appendix~\ref{app:dag}. These constraints define the executable planning space in DisasterBench: a workflow is considered valid only when tool
selection, parameter bindings, and inter-step dependencies remain
consistent across stages. DisasterBench uses a single global workflow structure shared across all tasks, reflecting the reusable nature of disaster-response tool ecosystems. The same tool may participate in multiple workflows, and the same intermediate output may support different downstream analyses depending on the response objective. Models are provided only with tool specifications and interface schemas, requiring them to infer executable workflows from task context alone without exposure to the underlying workflow structure itself.

\subsection{Task Generation}
\label{sec:task_generation}
Each task in DisasterBench is grounded in a valid executable workflow
derived from the workflow execution constraints and transformed into a
naturalistic planning scenario through an LLM-assisted pipeline with
human verification. A workflow specifies the sequence of tool calls,
parameter bindings, and inter-step dependencies required to complete the
task; workflow lengths range from 1 to 9 steps.

\textbf{Workflow Sampling.}
To capture the compositional diversity of disaster-response workflows,
we categorize tasks into three compositional settings:
\textit{Node} (single-step tasks invoking one tool),
\textit{Chain} (multi-step sequential workflows), and
\textit{Branching} (multi-step workflows containing one-to-many
dependencies). The benchmark contains 233 tasks in total
(Node:~35, Chain:~166, Branching:~32). Compositional difficulty
increases with workflow depth: Node tasks primarily test tool
grounding, while Chain and Branching tasks additionally require
parameter propagation and dependency consistency across multiple stages.
Detailed statistics are provided in Appendix~\ref{app:dataset_stats}.

\textbf{Description Synthesis.}
Task descriptions are generated through a self-instruct
procedure~\citep{wang2022self} using GPT-4o, deliberately reflecting
user intent without exposing the underlying execution constraints.
Descriptions are designed to prevent models from exploiting
surface-level patterns rather than inferring executable workflows
compositionally. Prompts are provided in Appendix~\ref{app:prompts}.

\textbf{Human Verification.}
All task--workflow pairs undergo two rounds of expert review by
annotators with backgrounds in remote sensing and disaster management.
Annotators verify semantic alignment between task descriptions and
ground-truth workflows, ensure that descriptions do not leak structural
cues such as tool names or intermediate representations, and confirm
that workflows remain valid under the execution constraints. Ambiguous
or inconsistent instances are revised or discarded, yielding a dataset
with expert-verified one-to-one alignment between task descriptions and
executable workflows.

\subsection{Evaluation Protocol}
\label{sec:eval_protocol}

\textbf{Strict Exact Match.}
We evaluate model-generated plans against ground truth using a strict
exact-match (EM) criterion. Given a predicted plan
$\hat{P} = (\hat{s}_1, \dots, \hat{s}_{\hat{T}})$ and a ground-truth
plan $P = (s_1, \dots, s_T)$, where each step $s_t$ specifies the agent
identity and its execution-consistent bindings and dependencies, a
prediction is correct if and only if:
\begin{equation}
\label{eq:em}
\text{EM}(\hat{P}, P) =
\mathbb{I}(\hat{T}=T \land \forall t,\ \hat{s}_t=s_t).
\end{equation}
Step equality $\hat{s}_t=s_t$ is defined jointly over tool identity,
parameter bindings, dependency structure, and dependency content. This
strict criterion is necessary because even a single incorrect binding or
dependency reference may route downstream tools to different intermediate
outputs, obscuring the workflow-grounding failure that DisasterBench is
designed to diagnose. Step order is likewise non-interchangeable because
downstream steps consume outputs produced by specific upstream tools.

\textbf{Partial-Credit Diagnostic Metrics.}
While EM captures end-to-end correctness, it does not reveal which
dimension of a plan failed. We complement EM with three plan-level
partial-credit metrics, each a task-level boolean averaged over all
tasks: \textit{Tool Accuracy} (step indices and agent identities
match), \textit{Parameter Accuracy} (tool conditions plus matching
input/output bindings at every step), and \textit{Dependency Accuracy}
(tool conditions plus matching inter-step dependencies and references
at every step). By construction, both Parameter and Dependency Accuracy
imply Tool Accuracy, and EM implies all three; however, Parameter and
Dependency Accuracy are not mutually comparable, since they isolate
disjoint plan dimensions: parameter binding and inter-step dependency
propagation, respectively. The gaps between these metrics quantify the
separation between semantically plausible tool selection and
execution-consistent workflow generation, a diagnostic property we term
the \textit{semantic-execution gap}.

\textbf{First-Point-of-Failure (FPoF).}
To attribute failures at the step level, we propose
First-Point-of-Failure (FPoF), which localizes the earliest
divergence between a predicted and ground-truth plan. Each first failure
is classified into three categories: \textit{tool mismatch} (incorrect
tool grounding), \textit{Parameter Binding Error} (incorrect inputs,
outputs, or upstream references), and \textit{Structural Error} (failure
to produce an executable plan, including hallucinated steps, premature
termination, or format violations). By isolating the earliest root
cause, FPoF separates primary execution failures from cascaded downstream
errors. Sub-category definitions and per-method distributions are
provided in Section~\ref{sec:failure_modes} and
Appendix~\ref{app:eval}.

\begin{table*}[t]
\centering
\small
\setlength{\tabcolsep}{4.5pt}
\renewcommand{\arraystretch}{0.97}
\begin{tabular*}{\textwidth}{@{\extracolsep{\fill}}llccccc}
\toprule
\textbf{Category} & \textbf{Model} & \textbf{DP} & \textbf{CoT} & \textbf{ToT} &
\textbf{RAP} & \textbf{ReAct} \\
\midrule
\multirow{2}{*}{Frontier}
& Gemini 3.1 Pro Preview
  & 68.24 & 72.53 & \underline{\textbf{73.39}} & 72.10 & 69.96 \\
& GPT-5.4
  & 54.08 & 62.23 & 62.23 & \textbf{65.67} & 64.38 \\
\midrule
\multirow{7}{*}{\shortstack[l]{Strong\\Open-source}}
& DeepSeek-V3.2
  & 57.94 & 36.05 & 50.21 & 34.33 & \textbf{61.80} \\
& DeepSeek-R1
  & 54.08 & \textbf{62.66} & 53.22 & 35.19 & 56.22 \\
& Qwen3-Max
  & 55.79 & 62.66 & \textbf{63.95} & 63.09 & 62.23 \\
& Qwen3-Max-Thinking
  & 57.51 & \textbf{66.52} & 62.23 & 61.37 & 61.37 \\
& Qwen3.5-27B
  & 58.37 & \textbf{62.23} & 59.23 & 58.80 & 59.23 \\
& Gemma-4-31B
  & \textbf{66.52} & 63.09 & 58.37 & 50.64 & 62.23 \\
& Llama-3.3-70B
  & 38.63 & \textbf{58.80} & 41.63 & 28.76 & 38.63 \\
\midrule
\multirow{5}{*}{\shortstack[l]{Efficiency-\\oriented}}
& Qwen3.5-9B
  & 45.92 & 61.37 & 59.23 & 42.92 & \textbf{65.67} \\
& Llama-3.1-8B
  & 15.88 & 26.18 & 16.31 & \textbf{28.33} & 26.61 \\
& Ministral-14B
  & 20.17 & 33.48 & 16.31 & 18.88 & \textbf{50.64} \\
& Ministral-8B
  & 25.32 & 31.33 & 21.03 & 12.88 & \textbf{44.21} \\
& Ministral-3B
  & 10.30 & 21.03 & 18.88 &  3.86 & \textbf{36.91} \\
\midrule
\multirow{4}{*}{Baselines}
& Random
  & \multicolumn{5}{c}{0.00 \ (Agent Acc.: $\sim$0\%)} \\
& DAG-Greedy (TF-IDF)
  & \multicolumn{5}{c}{0.00 \ (Agent Acc.: 5.58\%)} \\
& Shortest Path
  & \multicolumn{5}{c}{0.00 \ (Agent Acc.: 51.07\%)} \\
& Oracle-Random
  & \multicolumn{5}{c}{0.00 \ (Agent Acc.: 100\%)} \\
\midrule
\multicolumn{2}{l}{\textit{Average (LLMs)}}
  & 44.91 & 51.44 & 46.87 & 41.20 & 54.29 \\
\bottomrule
\end{tabular*}
\caption{Exact-match accuracy (\%) of 14 LLMs across five planning methods on
DisasterBench. \textbf{Bold} indicates the best result per model. The overall best
result is \underline{underlined}. Structural baselines do not distinguish planning
methods and are reported as single EM scores.}
\label{tab:main_results}
\normalsize
\renewcommand{\arraystretch}{1.0}
\end{table*}


\section{Benchmark Evaluation}

Our evaluation is designed not only to rank models, but also to diagnose
where executable workflow grounding fails. We therefore report
exact-match accuracy together with baselines, partial-credit
metrics, and First-Point-of-Failure analyses, separating semantic
tool-grounding failures from execution-level parameter binding,
dependency consistency, and structural generation errors.

\subsection{Experimental Setup}
\label{sec:experimental_setup}

\noindent\textbf{Models.}
We evaluate 14 large language models spanning a wide range of scale,
training recipe, and reasoning specialization, grouped into three tiers:
\textit{Frontier} proprietary models, \textit{Strong open-source} models
from competitive open-weight families and reasoning-optimized variants,
and \textit{Efficiency-oriented} models covering smaller Llama/Qwen
variants and the Ministral family. The full model list with category
assignments is shown in Table~\ref{tab:main_results}; complete
identifiers, API endpoints, and access dates are provided in
Appendix~\ref{app:models}. This selection enables controlled comparisons
across model capacity and reasoning specialization in executable
workflow grounding. Complete implementation details and hyperparameter configurations are provided in Appendix~\ref{app:implementation}.

\noindent\textbf{Planning Methods.}
We evaluate five representative planning strategies.

\textit{Direct Prompting (DP):}
Generates the complete plan in a single forward pass.

\textit{Chain-of-Thought (CoT)}~\citep{wei2022chain}:
Elicits intermediate reasoning before generating the final plan.

\textit{Tree-of-Thought (ToT)}~\citep{yao2023tree}:
Performs breadth-first search over candidate plans with LLM-based pruning.

\textit{Reasoning via Planning (RAP)}~\citep{hao2023reasoning}:
Uses Monte Carlo Tree Search guided by usefulness and self-consistency rewards.

\textit{ReAct}~\citep{yao2022react}:
Interleaves reasoning and tool selection over the current workflow state.

\noindent\textbf{Structural Baselines.}
We include four non-LLM baselines to isolate the contribution of
language-grounded workflow generation; all achieve $0\%$ EM
regardless of how much structural information they receive,
confirming that execution-consistent workflow generation requires
language-grounded parameter binding beyond tool selection alone
(Appendix~\ref{app:baselines}).

\noindent\textbf{Implementation.}
All models are queried via official APIs with prompt-level JSON schema
enforcement only; unparseable outputs are treated as incorrect.
Full configurations are provided in Appendix~\ref{app:implementation}.


\subsection{Main Results}
\label{sec:main_results}

Table~\ref{tab:main_results} reports exact-match (EM) accuracy for all
14 models across five planning methods. The results reveal that planning
strategy and model capacity interact in non-uniform ways: no single
planning strategy is universally optimal, and method effectiveness
depends systematically on model tier and workflow-grounding performance.

\textbf{Finding 1: Model capacity is the primary bottleneck.}
The performance gap between frontier and efficiency-oriented models
substantially exceeds the variation introduced by any individual
planning strategy, suggesting that maintaining execution consistency
across multi-step workflows is fundamentally constrained by model
capacity rather than reasoning paradigm alone.

\textbf{Finding 2: Method optimality is tier-dependent.}
Frontier models peak under search-based methods (Gemini under ToT,
$73.39\%$; GPT-5.4 under RAP, $65.67\%$), while strong open-source
models most often benefit from CoT. Efficiency-oriented models almost
uniformly favor ReAct, with gains of up to $+17$ percentage points (pp) over CoT
(Ministral-14B: $50.64\%$ vs.\ $33.48\%$). This suggests that global
search-based reasoning is reliably exploitable only by sufficiently
capable models, while local interleaved reasoning is more robust when
execution consistency becomes difficult to maintain across steps.

\textbf{Finding 3: Search-based methods require reliable
self-evaluation.}
RAP is competitive on frontier models (Gemini: $72.10\%$, GPT-5.4:
$65.67\%$) but degrades sharply on weaker models (Llama-3.1-8B:
$28.33\%$, Ministral-3B: $3.86\%$), even underperforming DP for
several efficiency-tier models. This pattern suggests that weaker
models struggle to reliably evaluate partially generated workflows,
causing MCTS search to amplify rather than correct execution
errors, a hypothesis we examine further in
Section~\ref{sec:failure_modes}.

\begin{table}[t]
\centering
\scriptsize
\setlength{\tabcolsep}{2pt}
\renewcommand{\arraystretch}{0.96}
\begin{tabular*}{\columnwidth}{@{\extracolsep{\fill}}>{\raggedright\arraybackslash}p{0.34\columnwidth}rrrrr@{}}
\toprule
\textbf{Error Type} & \textbf{DP} & \textbf{CoT} & \textbf{ToT} &
\textbf{RAP} & \textbf{ReAct} \\
\midrule
\textbf{tool mismatch}              & 50.4 & 54.6 & 52.3 & 32.2 & 58.2 \\
\midrule
\textbf{Parameter Binding}           & 34.6 & 31.7 & 31.0 & 50.1 & 28.7 \\
\quad\textit{Parameter}              & 34.5 & 31.5 & 29.3 & 48.5 & 28.7 \\
\quad\textit{Dependency}             &  0.0 &  0.2 &  1.7 &  1.2 &  0.0 \\
\quad\textit{Dep.-content}           &  0.1 &  0.0 &  0.1 &  0.4 &  0.0 \\
\midrule
\textbf{Structural}                  & 15.0 & 13.7 & 16.7 & 17.7 & 13.1 \\
\quad\textit{Halluc. steps}          & 13.0 & 11.9 & 15.6 &  6.6 & 11.3 \\
\quad\textit{Early stop}             &  1.9 &  1.8 &  1.1 & 11.1 &  1.6 \\
\quad\textit{Format}                 &  0.1 &  0.0 &  0.0 &  0.0 &  0.2 \\
\midrule
Failures ($n$)                       & 1797 & 1584 & 1733 & 1918 & 1491 \\
\bottomrule
\end{tabular*}
\caption{FPoF error distribution by planning method (\% of method
failures). Fine-grained sub-categories shown in italics.
Columns sum to $100\%$ within each method.}
\label{tab:fpof}
\end{table}

\subsection{Semantic-Execution Gap}
\label{sec:gap}

We define the \textbf{semantic-execution gap} as the per-task difference
between Tool Accuracy (A) and exact-match accuracy (EM) on the same
prediction. A positive gap indicates that the model selected the correct
tool sequence but failed to produce a fully executable plan along at
least one parameter-binding or dependency-propagation requirement. Full
results are reported in Table~\ref{tab:partial_credit}
(Appendix~\ref{app:semantic_gap}); the gap is present for every model
under every method, with average values ranging from ${\sim}5$ to
${\sim}11$pp across all model--method combinations, confirming that
\emph{semantically plausible tool selection is consistently easier than
execution-consistent workflow grounding}.

\textbf{The gap is more strongly associated with planning
method than with model tier.}
Averaged across all models, the gap varies systematically
with planning method (DP: $8.2$pp, CoT: $6.7$pp, ToT:
$9.0$pp, RAP: $7.9$pp, ReAct: $6.8$pp) but does not
cleanly separate model tiers. For example, Gemini's
average gap ($9.3$pp) is comparable to that of
Ministral-3B ($9.7$pp). Methods with explicit step-by-step
reasoning, such as CoT and ReAct, produce the smallest
gaps, suggesting that articulated intermediate reasoning
helps preserve local execution consistency even when global
workflow generation remains error-prone. Conversely,
single-pass generation and global search-based paradigms
yield larger gaps, consistent with their greater
susceptibility to execution-level failures.

\textbf{Parameter binding is an independent execution bottleneck.}
The largest gap occurs for Llama-3.3-70B under DP, which selects
correct tools on $60.94\%$ of tasks but achieves only $38.63\%$
EM, corresponding to a $22.31$pp gap. Even when tool grounding
succeeds, execution-consistent parameter binding and dependency
propagation remain distinct failure sources, confirming that correct
tool sequences are necessary but not sufficient for executable
workflow grounding. The DeepSeek-V3.2 CoT collapse, by contrast, is not
associated with a widening of this gap but with a uniform drop in both
A and EM; we treat it separately as an instance of
\emph{instruction clash} in Section~\ref{sec:instruction_clash}.

These results suggest that workflow grounding introduces
execution-consistency requirements beyond semantic tool selection alone:
Models must not only identify plausible tools, but also maintain valid
parameter bindings and dependency propagation throughout the workflow.

\subsection{Depth Analysis}
\label{sec:depth}

Figure~\ref{fig:depth_accuracy} plots exact-match accuracy as a function
of workflow depth, micro-averaged across all models for each planning
method. Performance degrades sharply as workflow depth increases across
all methods and model tiers: even the strongest model, Gemini~3.1 Pro
Preview, falls from near-perfect accuracy at depth~1 to roughly
$53$--$61\%$ at depth~4 across methods, confirming that deeper workflows
introduce a difficulty dimension not eliminated by increased model
capacity alone. Method differences also narrow with depth, suggesting
that accumulated execution errors eventually overwhelm the benefits of
individual reasoning strategies. This convergence is consistent with the
view that deeper workflows impose compounding
execution-consistency requirements rather than simply increased
reasoning complexity. We examine this pattern further through failure
attribution in Section~\ref{sec:failure_modes}.

\subsection{Failure Mode Analysis}
\label{sec:failure_modes}

Table~\ref{tab:fpof} reports the FPoF error distribution aggregated
over all 70 model--method combinations and 8{,}523 total failures.
Because FPoF records the earliest divergence from the ground-truth
workflow, these distributions reflect likely root causes rather than
downstream cascading symptoms. Tool mismatch and Parameter Binding
Error together account for over $80\%$ of failures under every method,
confirming that executable workflow grounding failures dominate over
low-level structural generation failures.

\textbf{Tool mismatch is the dominant semantic grounding failure.}
For four of the five methods, tool mismatch is the most frequent first
error ($50.4\%$--$58.2\%$), reflecting the difficulty of semantic
discrimination among functionally overlapping but operationally distinct
tools. RAP is a qualitative exception: under RAP, Parameter Binding
Error ($50.1\%$) overtakes tool mismatch ($32.2\%$), suggesting that
MCTS often reaches a plausible initial tool but fails to maintain
execution-consistent parameter bindings and dependency propagation in
subsequent steps.

\textbf{RAP exhibits a distinctive early-termination signature.}
RAP shows substantially elevated Early Stop failures ($11.1\%$
vs.\ $1$--$2\%$ elsewhere) and suppressed Hallucinated Steps ($6.6\%$
vs.\ $11$--$16\%$), pointing to a search dynamic that truncates
workflows prematurely. This is consistent with the view that weaker
models cannot reliably evaluate partially generated workflows, causing
local execution uncertainty to accumulate during search expansion and
MCTS to terminate before reaching a complete executable workflow.

\textbf{Structural errors reflect semantic over-generation, not
formatting failures.}
Structural Errors account for $13$--$18\%$ of failures but are
dominated by Hallucinated Steps rather than format violations
($\leq 0.2\%$ across methods), indicating that low-level formatting
is rarely the primary failure source. Instead, models tend toward
semantic over-generation: inserting semantically plausible but
unwarranted tool calls beyond the ground-truth workflow. Per-model
FPoF breakdowns are provided in Appendix~\ref{app:fpof}.

\subsection{Reasoning-Optimized Models and Instruction Clash}
\label{sec:instruction_clash}

Verbose intermediate reasoning does not uniformly improve executable
workflow generation. In some configurations it helps; in others it
actively disrupts structured output production, an effect that depends
on both the planning paradigm and the model's reasoning style.

\textbf{Instruction clash: when reasoning disrupts executable
generation.}
The most striking case is DeepSeek-V3.2 under CoT, where EM collapses
from $57.94\%$ (DP) to $36.05\%$, a drop of $21.89$pp. The FPoF
distribution reveals a sharp redistribution rather than a uniform
decline: under DP, only $32.7\%$ of failures are Parameter Errors, but
under CoT this share jumps to $76.5\%$ ($+43.8$pp). This sharp
redistribution, from tool-grounding failures under DP to
parameter-binding failures under CoT, indicates that the verbose
reasoning trace actively interferes with schema-consistent structured
output generation rather than simply degrading tool selection. We term
this \textbf{instruction clash}: CoT introduces competing generation
objectives, namely free-form intermediate reasoning and
schema-constrained executable workflow generation. For some models, the
reasoning objective dominates, destabilizing parameter binding and
dependency propagation in the final structured output.

\textbf{The effect is paradigm-dependent.}
Reasoning-optimized variants help under CoT, where explicit reasoning
aligns with the prompt's intent (Qwen3-Max-Thinking: $66.52\%$
vs.\ Qwen3-Max: $62.66\%$), but underperform base models under DP
(DeepSeek-R1: $54.08\%$ vs.\ DeepSeek-V3.2: $57.94\%$), where extended
reasoning does not improve executable grounding. Under RAP, the gap
between base and reasoning variants remains small (within ${\sim}2$pp),
suggesting that when an external search procedure orchestrates plan
construction, the model's internal reasoning style has limited
additional impact on execution consistency. An ablation externalizing format constraints via API-level JSON mode
confirms that the collapse is caused by formatting conflict rather than
degraded reasoning (Appendix~\ref{app:clash_ablation}).

\textbf{Implications.}
These results suggest that reasoning optimization does not uniformly
transfer to executable workflow grounding. The conflict between verbose
intermediate reasoning and schema-constrained structured generation
emerges as a measurable and diagnosable failure mode, highlighting the
importance of evaluating reasoning quality jointly with executable
output controllability rather than through aggregate accuracy alone.

\section{Conclusion}

We present DisasterBench, a benchmark for evaluating executable
workflow grounding in disaster-response multi-agent systems.
By requiring models to generate execution-consistent plans over
semantically similar but operationally distinct tools, DisasterBench
exposes failure modes that aggregate metrics alone cannot reveal.
We further propose First-Point-of-Failure (FPoF), a step-level
diagnostic framework that separates root causes from cascaded
downstream errors, enabling more precise failure attribution.

Across 14 LLMs and five planning paradigms, our evaluation reveals
that performance degrades sharply with workflow depth, tool mismatch
and parameter-binding failures dominate first errors, and verbose
intermediate reasoning can disrupt structured output generation via
instruction clash. Together, these findings suggest that strong
semantic reasoning alone is insufficient for reliable
execution-grounded planning, motivating future research on structured
workflow coordination in safety-critical domains.

\section{Limitations}

DisasterBench currently evaluates workflow grounding under predefined
agent specifications and benchmark settings. Future work may incorporate
interactive execution feedback and more dynamic environments to study
adaptive workflow planning under realistic deployment conditions.
Extending the benchmark to multilingual settings is another important
direction for future research.

\section*{Ethics Statement}
\label{sec:ethics}

DisasterBench is designed solely for research purposes to evaluate
LLM-based planning and coordination capabilities in structured
disaster-response workflows. The benchmark is constructed from publicly
available research artifacts, including open-source models, published
methodologies, and documented disaster-management pipelines.

The benchmark does not contain personally identifiable information (PII),
private user data, or sensitive operational records. All workflow tasks
and agent interfaces are abstracted at the system level and are intended
for evaluation of planning consistency rather than real-world emergency
deployment.

Although the benchmark focuses on disaster-response scenarios, the
evaluated systems should not be interpreted as deployment-ready decision
support tools. Incorrect workflow generation, parameter binding failures,
or dependency propagation errors may lead to unsafe or misleading outputs
in safety-critical environments. Therefore, benchmark results should be
treated as diagnostic indicators of current model limitations rather than
evidence of reliable autonomous disaster management capabilities.

We also acknowledge that advances in agentic planning systems may
introduce broader societal risks, including over-reliance on automated
decision-making and misuse in high-stakes settings without sufficient
human oversight. We encourage future deployment-oriented research to
incorporate domain experts, human-in-the-loop verification, and rigorous
safety validation procedures.

\section*{Acknowledgements}
This work used the ACES at Texas A\&M University, DeltaAI, and Delta GPU resources at the National Center for Supercomputing Applications through allocations CIV250019 and CIV250021 from the Advanced Cyberinfrastructure Coordination Ecosystem: Services \& Support (ACCESS) program, which is supported by U.S. National Science Foundation grants \#2138259, \#2138286, \#2138307, \#2137603, and \#2138296.
We also acknowledge the use of high-performance computing resources provided by the Texas A\&M University High Performance Research Computing (HPRC) facility.

\bibliography{custom}

@inproceedings{qin2024toolllm,
  title={Toolllm: Facilitating large language models to master 16000+ real-world apis},
  author={Qin, Yujia and Liang, Shihao and Ye, Yining and Zhu, Kunlun and Yan, Lan and Lu, Yaxi and Lin, Yankai and Cong, Xin and Tang, Xiangru and Qian, Bill and others},
  booktitle={International Conference on Learning Representations},
  volume={2024},
  pages={9695--9717},
  year={2024}
}

@article{patil2024gorilla,
  title={Gorilla: Large language model connected with massive apis},
  author={Patil, Shishir G and Zhang, Tianjun and Wang, Xin and Gonzalez, Joseph E},
  journal={Advances in Neural Information Processing Systems},
  volume={37},
  pages={126544--126565},
  year={2024}
}

@inproceedings{liu2024agentbench,
  title={Agentbench: Evaluating llms as agents},
  author={Liu, Xiao and Yu, Hao and Zhang, Hanchen and Xu, Yifan and Lei, Xuanyu and Lai, Hanyu and Gu, Yu and Ding, Hangliang and Men, Kaiwen and Yang, Kejuan and others},
  booktitle={International Conference on Learning Representations},
  volume={2024},
  pages={52989--53046},
  year={2024}
}

@article{shen2024taskbench,
  title={Taskbench: Benchmarking large language models for task automation},
  author={Shen, Yongliang and Song, Kaitao and Tan, Xu and Zhang, Wenqi and Ren, Kan and Yuan, Siyu and Lu, Weiming and Li, Dongsheng and Zhuang, Yueting},
  journal={Advances in Neural Information Processing Systems},
  volume={37},
  pages={4540--4574},
  year={2024}
}

@article{wei2022chain,
  title={Chain-of-thought prompting elicits reasoning in large language models},
  author={Wei, Jason and Wang, Xuezhi and Schuurmans, Dale and Bosma, Maarten and Xia, Fei and Chi, Ed and Le, Quoc V and Zhou, Denny and others},
  journal={Advances in neural information processing systems},
  volume={35},
  pages={24824--24837},
  year={2022}
}

@article{yao2023tree,
  title={Tree of thoughts: Deliberate problem solving with large language models},
  author={Yao, Shunyu and Yu, Dian and Zhao, Jeffrey and Shafran, Izhak and Griffiths, Tom and Cao, Yuan and Narasimhan, Karthik},
  journal={Advances in neural information processing systems},
  volume={36},
  pages={11809--11822},
  year={2023}
}

@article{yao2022react,
  title={React: Synergizing reasoning and acting in language models},
  author={Yao, Shunyu and Zhao, Jeffrey and Yu, Dian and Du, Nan and Shafran, Izhak and Narasimhan, Karthik and Cao, Yuan},
  journal={arXiv preprint arXiv:2210.03629},
  year={2022}
}

@inproceedings{hao2023reasoning,
  title={Reasoning with language model is planning with world model},
  author={Hao, Shibo and Gu, Yi and Ma, Haodi and Hong, Joshua and Wang, Zhen and Wang, Daisy and Hu, Zhiting},
  booktitle={Proceedings of the 2023 Conference on Empirical Methods in Natural Language Processing},
  pages={8154--8173},
  year={2023}
}

@article{algiriyage2022multi,
  title={Multi-source multimodal data and deep learning for disaster response: a systematic review},
  author={Algiriyage, Nilani and Prasanna, Raj and Stock, Kristin and Doyle, Emma EH and Johnston, David},
  journal={SN Computer Science},
  volume={3},
  number={1},
  pages={92},
  year={2022},
  publisher={Springer}
}

@article{bentivoglio2022deep,
  title={Deep learning methods for flood mapping: a review of existing applications and future research directions},
  author={Bentivoglio, Roberto and Isufi, Elvin and Jonkman, Sebastian Nicolaas and Taormina, Riccardo},
  journal={Hydrology and Earth System Sciences Discussions},
  volume={2022},
  pages={1--50},
  year={2022},
  publisher={G{\"o}ttingen, Germany}
}

@article{gupta2019xbd,
  title={xbd: A dataset for assessing building damage from satellite imagery},
  author={Gupta, Ritwik and Hosfelt, Richard and Sajeev, Sandra and Patel, Nirav and Goodman, Bryce and Doshi, Jigar and Heim, Eric and Choset, Howie and Gaston, Matthew},
  journal={arXiv preprint arXiv:1911.09296},
  year={2019}
}

@article{wang2022self,
  title={Self-consistency improves chain of thought reasoning in language models},
  author={Wang, Xuezhi and Wei, Jason and Schuurmans, Dale and Le, Quoc and Chi, Ed and Narang, Sharan and Chowdhery, Aakanksha and Zhou, Denny},
  journal={arXiv preprint arXiv:2203.11171},
  year={2022}
}

@article{gebru2021datasheets,
  title={Datasheets for datasets},
  author={Gebru, Timnit and Morgenstern, Jamie and Vecchione, Briana and Vaughan, Jennifer Wortman and Wallach, Hanna and Iii, Hal Daum{\'e} and Crawford, Kate},
  journal={Communications of the ACM},
  volume={64},
  number={12},
  pages={86--92},
  year={2021},
  publisher={ACM New York, NY, USA}
}

@inproceedings{yin2017syntactic,
  title={A syntactic neural model for general-purpose code generation},
  author={Yin, Pengcheng and Neubig, Graham},
  booktitle={Proceedings of the 55th Annual Meeting of the Association for Computational Linguistics (Volume 1: Long Papers)},
  pages={440--450},
  year={2017}
}

@inproceedings{scholak2021picard,
  title={PICARD: Parsing incrementally for constrained auto-regressive decoding from language models},
  author={Scholak, Torsten and Schucher, Nathan and Bahdanau, Dzmitry},
  booktitle={Proceedings of the 2021 conference on empirical methods in natural language processing},
  pages={9895--9901},
  year={2021}
}

@inproceedings{shin2021constrained,
  title={Constrained language models yield few-shot semantic parsers},
  author={Shin, Richard and Lin, Christopher and Thomson, Sam and Chen Jr, Charles and Roy, Subhro and Platanios, Emmanouil Antonios and Pauls, Adam and Klein, Dan and Eisner, Jason and Van Durme, Benjamin},
  booktitle={Proceedings of the 2021 conference on empirical methods in natural language processing},
  pages={7699--7715},
  year={2021}
}

@article{chen2026disastqa,
  title={DisastQA: A Comprehensive Benchmark for Evaluating Question Answering in Disaster Management},
  author={Chen, Zhitong and Yin, Kai and Dong, Xiangjue and Liu, Chengkai and Li, Xiangpeng and Xiao, Yiming and Li, Bo and Ma, Junwei and Mostafavi, Ali and Caverlee, James},
  journal={arXiv preprint arXiv:2601.03670},
  year={2026}
}

@article{liu2025floodsql,
  title={FloodSQL-Bench: A Retrieval-Augmented Benchmark for Geospatially-Grounded Text-to-SQL},
  author={Liu, Hanzhou and Yin, Kai and Chen, Zhitong and Liu, Chenyue and Mostafavi, Ali},
  journal={arXiv preprint arXiv:2512.12084},
  year={2025}
}

@article{tran2025multiagent,
  title={Multi-agent collaboration mechanisms: A survey of llms},
  author={Tran, Khanh-Tung and Dao, Dung and Nguyen, Minh-Duong and Pham, Quoc-Viet and O'Sullivan, Barry and Nguyen, Hoang D},
  journal={arXiv preprint arXiv:2501.06322},
  year={2025}
}

@inproceedings{khanna2024diffusionsat,
  title={Diffusionsat: A generative foundation model for satellite imagery},
  author={Khanna, Samar and Liu, Patrick and Zhou, Linqi and Meng, Chenlin and Rombach, Robin and Burke, Marshall and Lobell, David and Ermon, Stefano},
  booktitle={International Conference on Learning Representations},
  volume={2024},
  pages={5586--5604},
  year={2024}
}

@inproceedings{patil2023multi,
  title={Multi-weather image restoration via domain translation},
  author={Patil, Prashant W and Gupta, Sunil and Rana, Santu and Venkatesh, Svetha and Murala, Subrahmanyam},
  booktitle={Proceedings of the IEEE/CVF International Conference on Computer Vision},
  pages={21696--21705},
  year={2023}
}

@article{cong2022satmae,
  title={Satmae: Pre-training transformers for temporal and multi-spectral satellite imagery},
  author={Cong, Yezhen and Khanna, Samar and Meng, Chenlin and Liu, Patrick and Rozi, Erik and He, Yutong and Burke, Marshall and Lobell, David and Ermon, Stefano},
  journal={Advances in Neural Information Processing Systems},
  volume={35},
  pages={197--211},
  year={2022}
}

@inproceedings{li2023anomaly,
  title={Anomaly segmentation for high-resolution remote sensing images based on pixel descriptors},
  author={Li, Jingtao and Wang, Xinyu and Zhao, Hengwei and Wang, Shaoyu and Zhong, Yanfei},
  booktitle={Proceedings of the AAAI Conference on Artificial Intelligence},
  volume={37},
  number={4},
  pages={4426--4434},
  year={2023}
}

@inproceedings{kuckreja2024geochat,
  title={Geochat: Grounded large vision-language model for remote sensing},
  author={Kuckreja, Kartik and Danish, Muhammad Sohail and Naseer, Muzammal and Das, Abhijit and Khan, Salman and Khan, Fahad Shahbaz},
  booktitle={Proceedings of the IEEE/CVF conference on computer vision and pattern recognition},
  pages={27831--27840},
  year={2024}
}

@inproceedings{li2023geolm,
  title={Geolm: Empowering language models for geospatially grounded language understanding},
  author={Li, Zekun and Zhou, Wenxuan and Chiang, Yao-Yi and Chen, Muhao},
  booktitle={Proceedings of the 2023 conference on empirical methods in natural language processing},
  pages={5227--5240},
  year={2023}
}

@inproceedings{zheng2020foreground,
  title={Foreground-aware relation network for geospatial object segmentation in high spatial resolution remote sensing imagery},
  author={Zheng, Zhuo and Zhong, Yanfei and Wang, Junjue and Ma, Ailong},
  booktitle={Proceedings of the IEEE/CVF conference on computer vision and pattern recognition},
  pages={4096--4105},
  year={2020}
}

@inproceedings{yu2024diffcast,
  title={Diffcast: A unified framework via residual diffusion for precipitation nowcasting},
  author={Yu, Demin and Li, Xutao and Ye, Yunming and Zhang, Baoquan and Luo, Chuyao and Dai, Kuai and Wang, Rui and Chen, Xunlai},
  booktitle={Proceedings of the IEEE/CVF Conference on Computer Vision and Pattern Recognition},
  pages={27758--27767},
  year={2024}
}

@article{kaur2023large,
  title={Large-scale building damage assessment using a novel hierarchical transformer architecture on satellite images},
  author={Kaur, Navjot and Lee, Cheng-Chun and Mostafavi, Ali and Mahdavi-Amiri, Ali},
  journal={Computer-Aided Civil and Infrastructure Engineering},
  volume={38},
  number={15},
  pages={2072--2091},
  year={2023},
  publisher={Wiley Online Library}
}

@inproceedings{wu2024vadclip,
  title={Vadclip: Adapting vision-language models for weakly supervised video anomaly detection},
  author={Wu, Peng and Zhou, Xuerong and Pang, Guansong and Zhou, Lingru and Yan, Qingsen and Wang, Peng and Zhang, Yanning},
  booktitle={Proceedings of the AAAI conference on artificial intelligence},
  volume={38},
  number={6},
  pages={6074--6082},
  year={2024}
}

@inproceedings{wang2022event,
  title={Event-aware multimodal mobility nowcasting},
  author={Wang, Zhaonan and Jiang, Renhe and Xue, Hao and Salim, Flora D and Song, Xuan and Shibasaki, Ryosuke},
  booktitle={Proceedings of the AAAI Conference on Artificial Intelligence},
  volume={36},
  number={4},
  pages={4228--4236},
  year={2022}
}

@article{huang2023counting,
  title={Counting crowds in bad weather},
  author={Huang, Zhi-Kai and Chen, Wei-Ting and Chiang, Yuan-Chun and Kuo, Sy-Yen and Yang, Ming-Hsuan},
  journal={arXiv preprint arXiv:2306.01209},
  year={2023}
}

@inproceedings{chen2023saras,
  title={SARAS-Net: Scale and relation aware Siamese network for change detection},
  author={Chen, Chao-Peng and Hsieh, Jun-Wei and Chen, Ping-Yang and Hsieh, Yi-Kuan and Wang, Bor-Shiun},
  booktitle={Proceedings of the AAAI Conference on Artificial Intelligence},
  volume={37},
  number={12},
  pages={14187--14195},
  year={2023}
}

@inproceedings{liu2022image,
  title={Image-adaptive YOLO for object detection in adverse weather conditions},
  author={Liu, Wenyu and Ren, Gaofeng and Yu, Runsheng and Guo, Shi and Zhu, Jianke and Zhang, Lei},
  booktitle={Proceedings of the AAAI conference on artificial intelligence},
  volume={36},
  number={2},
  pages={1792--1800},
  year={2022}
}

@article{yin2023integrated,
  title={An integrated resilience assessment model of urban transportation network: A case study of 40 cities in China},
  author={Yin, Kai and Wu, Jianjun and Wang, Weiping and Lee, Der-Horng and Wei, Yun},
  journal={Transportation Research Part A: Policy and Practice},
  volume={173},
  pages={103687},
  year={2023},
  publisher={Elsevier}
}

@inproceedings{kazadi2024pluvial,
  title={Pluvial flood emulation with hydraulics-informed message passing},
  author={Kazadi, Arnold and Doss-Gollin, James and Da Silva, Arlei Lopes},
  booktitle={Forty-First International Conference on Machine Learning},
  year={2024}
}

\clearpage
\appendix

\section{Model and Tool Details}
\label{app:models}

\subsection{LLM Identifiers and Access Information}

Table~\ref{tab:model_ids} lists the complete model identifiers, API
providers, and access dates for all 14 LLMs evaluated in this work.

\begin{table*}[t]
\centering
\footnotesize
\setlength{\tabcolsep}{4pt}
\begin{tabular*}{\textwidth}{@{\extracolsep{\fill}}p{0.24\textwidth}p{0.40\textwidth}p{0.14\textwidth}p{0.12\textwidth}}
\toprule
\textbf{Model (Paper Name)} & \textbf{API Identifier} & \textbf{Provider} & \textbf{Accessed} \\
\midrule
Gemini 3.1 Pro Preview   & \texttt{google/gemini-3.1-pro-preview}   & Google      & Apr.\ 2026 \\
GPT-5.4                  & \texttt{gpt-5.4}                         & OpenAI      & Apr.\ 2026 \\
\midrule
DeepSeek-V3.2            & \texttt{deepseek-ai/deepseek-v3.2}       & DeepSeek    & Apr.\ 2026 \\
DeepSeek-R1              & \texttt{deepseek-ai/deepseek-r1}         & DeepSeek    & Apr.\ 2026 \\
Qwen3-Max                & \texttt{qwen/qwen3-max}                  & Alibaba     & Apr.\ 2026 \\
Qwen3-Max-Thinking       & \texttt{qwen/qwen3-max-thinking}         & Alibaba     & Apr.\ 2026 \\
Qwen3.5-27B              & \texttt{qwen/qwen3.5-27b}                & Alibaba     & Apr.\ 2026 \\
Gemma-4-31B              & \texttt{google/gemma-4-31b}              & Google      & Apr.\ 2026 \\
Llama-3.3-70B            & \texttt{meta-llama/llama-3.3-70b}        & Meta        & Apr.\ 2026 \\
\midrule
Qwen3.5-9B               & \texttt{qwen/qwen3.5-9b}                 & Alibaba     & Mar.\ 2026 \\
Llama-3.1-8B             & \texttt{meta-llama/llama-3.1-8b}         & Meta        & Apr.\ 2026 \\
Ministral-14B            & \texttt{mistralai/ministral-14b}         & Mistral     & Apr.\ 2026 \\
Ministral-8B             & \texttt{mistralai/ministral-8b}          & Mistral     & Apr.\ 2026 \\
Ministral-3B             & \texttt{mistralai/ministral-3b}          & Mistral     & Apr.\ 2026 \\
\bottomrule
\end{tabular*}
\caption{Complete model identifiers, API providers, and access dates
for all 14 evaluated LLMs. All models are accessed via OpenRouter.}
\label{tab:model_ids}
\end{table*}

\subsection{Tool Source References}

Table~\ref{tab:agent_sources} lists the source publication or repository
for each of the 26 specialized tools in DisasterBench.

\begin{table*}[t]
\centering
\footnotesize
\setlength{\tabcolsep}{3.5pt}
\begin{tabular*}{\textwidth}{@{\extracolsep{\fill}}p{0.52\textwidth}p{0.42\textwidth}}
\toprule
\textbf{Tool} & \textbf{Source} \\
\midrule
\texttt{Temporal\_High\_Resolution\_Image\_Generation}   & \citet{khanna2024diffusionsat} \\
\texttt{High-Resolution\_Image\_Reconstructor}           & \citet{khanna2024diffusionsat} \\
\texttt{Metadata\_and\_Text\_Prompt\_Image\_Generation}  & \citet{khanna2024diffusionsat} \\
\texttt{Weather\_Degraded\_Image\_Restoration}           & \citet{patil2023multi} \\
\texttt{Temporal\_Image\_Sequence\_Classifier}           & \citet{cong2022satmae} \\
\texttt{RGB\_GeoImage\_Classifier}                       & \citet{cong2022satmae} \\
\texttt{Multi\_Spectral\_Classifier}                     & \citet{cong2022satmae}  \\
\texttt{Anomaly\_Detection\_Forest}                      & \citet{li2023anomaly} \\
\texttt{Landslide\_Segmentation}                         & \citet{li2023anomaly} \\
\texttt{Urban\_Anomaly\_Detection}                       & \citet{li2023anomaly} \\
\texttt{Change\_Mapping\_and\_Detection}                 & \citet{chen2023saras} \\
\texttt{Foggy\_Scenario\_Object\_Detection}              & \citet{liu2022image} \\
\texttt{Low-Light\_Object\_Detection}                    & \citet{liu2022image} \\
\texttt{Geospatial\_Object\_Segmentation}                & \citet{zheng2020foreground} \\
\texttt{Crowd\_Counting\_in\_Adverse\_Weather}           & \citet{huang2023counting} \\
\texttt{Building\_damage\_assessment}                    & \citet{kaur2023large} \\
\texttt{Precipitation\_Nowcasting}                       & \citet{yu2024diffcast} \\
\texttt{Flood\_depth\_prediction}                        &  \citet{kazadi2024pluvial} \\
\texttt{depth\_speed\_model}                             & \citet{yin2023integrated} \\
\texttt{Toponym\_Detection}                              & \citet{li2023geolm} \\
\texttt{Event\_detection}                                & \citet{li2023anomaly} \\
\texttt{Video\_anomaly\_detection}                       & \citet{wu2024vadclip} \\
\texttt{Post\_Disaster\_Mobility\_Recovery}              & \citet{wang2022event} \\
\texttt{Multimodal\_mobility\_prediction\_under\_events} & \citet{wang2022event} \\
\texttt{GeoChat}                                         & \citet{kuckreja2024geochat} \\
\texttt{precipitation\_data\_convert\_tool}              & \citet{yu2024diffcast}  \\
\bottomrule
\end{tabular*}
\caption{Source publications and repositories for all 26 tools in
DisasterBench.}
\label{tab:agent_sources}
\end{table*}
\section{Implementation Details}
\label{app:implementation}

\subsection{Shared Generation Parameters}

All models are queried via their respective official APIs.
The following generation parameters are shared across all planning methods
unless otherwise noted (Table~\ref{tab:shared_params}).

\begin{table}[h]
\centering
\scriptsize
\setlength{\tabcolsep}{2pt}
\begin{tabularx}{\columnwidth}{@{}>{\raggedright\arraybackslash}p{0.28\columnwidth}c>{\raggedright\arraybackslash}X@{}}
\toprule
\textbf{Parameter} & \textbf{Value} & \textbf{Notes} \\
\midrule
\texttt{max\_tokens}   & 8{,}192  & Per API call \\
\texttt{top\_p}        & 0.95     & Nucleus sampling \\
\texttt{top\_k}        & 20       & Where supported by API \\
\texttt{temperature}   & 0.0      & DP, CoT, ReAct (deterministic) \\
\texttt{temperature}   & 0.5      & ToT, RAP (stochastic search) \\
\bottomrule
\end{tabularx}
\caption{Shared generation hyperparameters across all models and methods.}
\label{tab:shared_params}
\end{table}

Temperature is overridden post-argument-parsing based on planning method:
deterministic decoding (\texttt{temperature}$=0.0$) is used for single-pass
methods (DP, CoT, ReAct), while a small degree of stochasticity
(\texttt{temperature}$=0.5$) is applied to search-based methods (ToT, RAP)
to enable exploration of diverse candidate trajectories.

\subsection{Reproducibility and Statistical Considerations}

For single-pass methods (DP, CoT, ReAct), we use deterministic decoding
(\texttt{temperature}$=0.0$), which produces identical outputs across
runs for a given model and prompt. Results for these methods are
therefore fully reproducible without variance. For search-based methods
(ToT, RAP), stochastic sampling (\texttt{temperature}$=0.5$) is
required to enable candidate diversity; results for these methods may
vary across runs. We report single-run results for all methods, which
is standard practice for large-scale LLM evaluation given the
substantial API cost of repeated runs across 14 models and 233 tasks.
Observed performance differences between methods within the same model
tier are generally large enough (often $>5$pp) to be robust to
run-to-run variance under these settings.

\subsection{Tree-of-Thought (ToT) Configuration}

We implement ToT as a two-step breadth-first search (BFS) with LLM-based
pruning. The search horizon is fixed at \textbf{2 expansion steps}, which
aligns with the average solution depth of the benchmark (2.53 steps).
At each step, candidate continuations are generated in parallel from all
active prefixes, scored by an LLM critic via majority voting, and pruned
to retain only the top-scoring prefix before the next expansion
(Table~\ref{tab:tot_params}).

\begin{table}[h]
\centering
\scriptsize
\setlength{\tabcolsep}{2pt}
\begin{tabularx}{\columnwidth}{@{}>{\raggedright\arraybackslash}p{0.30\columnwidth}c>{\raggedright\arraybackslash}X@{}}
\toprule
\textbf{Parameter} & \textbf{Value} & \textbf{Description} \\
\midrule
Branching factor  & 7
  & Candidate continuations generated per active prefix per step \\
Critic samples    & 5
  & Number of LLM critic calls per candidate for majority-vote scoring \\
Beam width        & 1
  & Number of prefixes retained after pruning at each step \\
Search depth      & 2
  & Number of BFS expansion steps (hardcoded) \\
Temperature       & 0.5
  & Applied to both generation and critic sampling \\
\bottomrule
\end{tabularx}
\caption{ToT hyperparameters (CLI defaults).}
\label{tab:tot_params}
\end{table}

The effective behaviour is a \emph{wide-expansion, narrow-retention} search:
at each step, up to $7 \times |\text{active prefixes}|$ candidates are
generated, each evaluated by 5 critic samples, and only the single
highest-scoring prefix is carried forward. This design prioritises
exploration at each step while controlling computational cost.

\subsection{Reasoning via Planning (RAP) Configuration}

RAP formulates plan generation as Monte Carlo Tree Search (MCTS) over the
tool pool. We use UCT-based node selection with a prior, aggregating child
rewards via mean and propagating them upward via max-child backpropagation.
The combined reward at each node is $R = r_0^\alpha \cdot r_1^{1-\alpha}$,
where $r_0$ is the usefulness reward and $r_1$ is the self-consistency reward
estimated by repeated sampling (Table~\ref{tab:rap_params}).

\begin{table}[h]
\centering
\scriptsize
\setlength{\tabcolsep}{2pt}
\renewcommand{\arraystretch}{0.96}
\begin{tabularx}{\columnwidth}{@{}>{\raggedright\arraybackslash}p{0.34\columnwidth}c>{\raggedright\arraybackslash}X@{}}
\toprule
\textbf{Parameter} & \textbf{Value} & \textbf{Description} \\
\midrule
Rollouts ($n_{\text{rollout}}$)
  & 5  & Number of MCTS simulation rollouts \\
Max tree depth
  & 5  & Maximum expansion depth (steps) \\
Sub-question samples ($n_{\text{subq}}$)
  & 16 & Candidate sub-steps sampled per node expansion \\
Confidence samples ($n_{\text{conf}}$)
  & 32 & Repeated samples for self-consistency reward $r_1$ \\
UCT exploration weight ($w$)
  & 1.0 & $w$ in UCT score $\bar{r} + w\sqrt{\ln N_p / N}$ \\
Reward exponent ($\alpha$)
  & 0.5 & Balances usefulness and self-consistency rewards \\
Default $r_1$
  & 1.0 & Fallback self-consistency reward when sampling fails \\
Reward aggregation
  & mean & Child reward aggregation in backpropagation \\
Temperature
  & 0.5 & Applied to all MCTS generation and sampling calls \\
\bottomrule
\end{tabularx}
\caption{RAP / MCTS hyperparameters (CLI defaults; per-model overrides
recorded in \texttt{results/<model>/rap/config.json}).}
\label{tab:rap_params}
\end{table}

\paragraph{Per-model configurations.}
For computationally expensive frontier models (e.g.,
\texttt{google/gemini-3.1-pro-preview}), we reduced \texttt{num\_rollouts}
and sampling counts to manage API cost while preserving the qualitative
search behaviour. Exact per-model configurations are recorded in
\texttt{results/<model>/rap/config.json} and are included in the released
codebase.

\paragraph{Note on \texttt{w\_exp} vs.\ \texttt{mcts\_exploration\_weight}.}
Some \texttt{config.json} files contain a field named
\texttt{mcts\_exploration\_weight} (e.g., 2.0); this field does
\emph{not} drive UCT exploration in our implementation---the effective
exploration weight is controlled exclusively by the \texttt{w\_exp}
argument passed directly to the \texttt{MCTS} constructor.

\subsection{Structured Output Enforcement}

All planning methods are prompted to produce output in a fixed JSON schema
specifying \texttt{step}, \texttt{agent}, \texttt{inputs}, \texttt{outputs},
\texttt{dependency}, and \texttt{dependency\_content} fields.
We deliberately do \emph{not} apply API-level schema enforcement (e.g.,
OpenAI function-calling, strict JSON mode, or post-hoc repair) beyond the
prompt-level specification. Outputs that cannot be parsed into the expected
schema are treated as incorrect predictions.

This design choice is intentional: API-level enforcement constrains the
generation space in ways that may artificially inflate performance, conflating
scaffolding capability with intrinsic planning ability. Investigating the
effect of schema enforcement on the instruction clash phenomenon---and whether
it mitigates CoT-induced formatting failures---is a promising direction for
future work.

\subsection{Token Budget and Computational Cost}

Our current implementation logs model outputs but does not instrument
\texttt{completion.usage} fields (prompt/completion token counts) in stored
prediction files. We acknowledge this as a limitation for strict
cost-controlled comparisons across methods, and plan to include token logging
in the public release.

As a qualitative guide, the five methods differ substantially in API call
count per task (Table~\ref{tab:call_counts}):

\begin{table}[h]
\centering
\scriptsize
\setlength{\tabcolsep}{2pt}
\renewcommand{\arraystretch}{0.96}
\begin{tabularx}{\columnwidth}{@{}>{\raggedright\arraybackslash}p{0.36\columnwidth}>{\raggedright\arraybackslash}X@{}}
\toprule
\textbf{Method} & \textbf{Approximate API calls per task} \\
\midrule
Direct Prompting (DP) & 1 \\
Chain-of-Thought (CoT) & 1 \\
ReAct & $\approx D$ (average $D = 2.53$ steps) \\
Tree-of-Thought (ToT) & $\approx 2 \times (n_{\text{gen}} + n_{\text{eval}}) = 2 \times (7 + 5) = 24$ \\
Reasoning via Planning (RAP) & $\approx n_{\text{rollout}} \times n_{\text{subq}} + n_{\text{conf}} \approx 50$--$200$ \\
\bottomrule
\end{tabularx}
\caption{Approximate API call count per task for each planning method.}
\label{tab:call_counts}
\end{table}

These figures reflect the default configurations; actual counts vary with
per-model overrides and early termination in MCTS.

\subsection{Code and Data Release}
\label{app:release}

The full benchmark---including the agent DAG (\texttt{graph\_desc.json}),
typed edge schema, 233 task JSON files with ground-truth plans, evaluation
code (including the FPoF analyzer and partial credit scorer), prompts for all
five planning methods, and per-model \texttt{config.json} files---will be
released publicly upon acceptance. Reproducibility scripts for all reported
experiments will be included. We are committed to enabling full reproduction
of every result reported in this paper.

\section{Complete Agent DAG Edge Table}
\label{app:dag}

Table~\ref{tab:dag_edges} lists all 81 directed typed edges in the
DisasterBench agent DAG, organized by subgraph family. To keep the table
compact, edge types are abbreviated as \mdedge{} (\texttt{model\_to\_data})
and \dmedge{} (\texttt{data\_to\_model}). Fan-in edges list multiple source
data nodes in brackets.

{
\setlength{\tabcolsep}{3pt}
\renewcommand{\arraystretch}{0.98}
\scriptsize

\begin{table*}[t]
\centering
\begin{tabularx}{\textwidth}{@{}>{\centering\arraybackslash}p{0.045\textwidth}>{\raggedright\arraybackslash}X>{\centering\arraybackslash}p{0.07\textwidth}>{\raggedright\arraybackslash}X@{}}
\toprule
\textbf{G.} & \textbf{Source} & \textbf{Type} & \textbf{Target} \\
\midrule

\multicolumn{4}{@{}l}{\textit{Subgraph I: Temporal High-Resolution Analysis (33 edges)}} \\

I & \dagcell{Metadata Prompt Image Generation} & \mdedge & \dagcell{Three Images Same Location} \\

I & \dagcell{Three Images Same Location} & \dmedge & \dagcell{Temporal HR Image Generation} \\

I & \dagcell{Temporal HR Image Generation} & \mdedge & \dagcell{HR Images Before and After Disaster} \\

I & \dagcell{HR Images Before and After Disaster} & \dmedge & \dagcell{Anomaly Detection Forest} \\

I & \dagcell{Anomaly Detection Forest} & \mdedge & \dagcell{Anomaly Map Forest} \\

I & \dagcell{Anomaly Map Forest} & \dmedge & \dagcell{GeoChat} \\

I & \dagcell{HR Images Before and After Disaster} & \dmedge & \dagcell{Landslide Segmentation} \\

I & \dagcell{Landslide Segmentation} & \mdedge & \dagcell{Landslide Map} \\

I & \dagcell{Landslide Map} & \dmedge & \dagcell{GeoChat} \\

I & \dagcell{HR Images Before and After Disaster} & \dmedge & \dagcell{Urban Anomaly Detection} \\

I & \dagcell{Urban Anomaly Detection} & \mdedge & \dagcell{Anomaly Map Urban} \\

I & \dagcell{Anomaly Map Urban} & \dmedge & \dagcell{GeoChat} \\

I & \dagcell{HR Images Before and After Disaster} & \dmedge & \dagcell{Change Mapping and Detection} \\

I & \dagcell{Change Mapping and Detection} & \mdedge & \dagcell{Change Map} \\

I & \dagcell{Change Map} & \dmedge & \dagcell{GeoChat} \\

I & \dagcell{HR Images Before and After Disaster} & \dmedge & \dagcell{Building Damage Assessment} \\

I & \dagcell{Building Damage Assessment} & \mdedge & \dagcell{Damage Classification Map} \\

I & \dagcell{Damage Classification Map} & \dmedge & \dagcell{GeoChat} \\

I & \dagcell{HR Images Before and After Disaster} & \dmedge & \dagcell{Temporal Image Sequence Classifier} \\

I & \dagcell{Temporal Image Sequence Classifier} & \mdedge & \dagcell{Temporal Classification Results} \\

I & \dagcell{Temporal Classification Results} & \dmedge & \dagcell{GeoChat} \\

I & \dagcell{HR Images Before and After Disaster} & \dmedge & \dagcell{Geospatial Object Segmentation} \\

I & \dagcell{Geospatial Object Segmentation} & \mdedge & \dagcell{Segmentation Map} \\

I & \dagcell{Segmentation Map} & \dmedge & \dagcell{GeoChat} \\

I & \dagcell{HR Images Before and After Disaster} & \dmedge & \dagcell{RGB GeoImage Classifier} \\

I & \dagcell{RGB GeoImage Classifier} & \mdedge & \dagcell{Predicted Category} \\

I & \dagcell{Predicted Category} & \dmedge & \dagcell{GeoChat} \\

I & \dagcell{GeoChat} & \mdedge & \dagcell{Contextual Description} \\

I & {[}\dagcell{Predicted Category} + \dagcell{Segmentation Map}{]} & \dmedge & \dagcell{GeoChat} \\

I & {[}\dagcell{Change Map} + \dagcell{Damage Classification Map}{]} & \dmedge & \dagcell{GeoChat} \\

I & {[}\dagcell{Damage Classification Map} + \dagcell{Temporal Classification Results}{]} & \dmedge & \dagcell{GeoChat} \\

I & {[}\dagcell{Damage Classification Map} + \dagcell{Temporal Classification Results} + \dagcell{Change Map}{]} & \dmedge & \dagcell{GeoChat} \\

\bottomrule
\end{tabularx}

\caption{Typed edges in the DisasterBench DAG by subgraph.}

\label{tab:dag_edges}

\end{table*}
}

{
\setlength{\tabcolsep}{3pt}
\renewcommand{\arraystretch}{0.98}
\scriptsize

\begin{table*}[t]
\centering
\begin{tabularx}{\textwidth}{@{}>{\centering\arraybackslash}p{0.045\textwidth}>{\raggedright\arraybackslash}X>{\centering\arraybackslash}p{0.07\textwidth}>{\raggedright\arraybackslash}X@{}}
\toprule
\textbf{G.} & \textbf{Source} & \textbf{Type} & \textbf{Target} \\
\midrule

\multicolumn{4}{@{}l}{\textit{Subgraph II: Image Reconstruction (23 edges)}} \\

II & \dagcell{Metadata Prompt Image Generation} & \mdedge & \dagcell{Low Resolution Multi-Spectral Image} \\

II & \dagcell{Low Resolution Multi-Spectral Image} & \dmedge & \dagcell{High Resolution Image Reconstructor} \\

II & \dagcell{High Resolution Image Reconstructor} & \mdedge & \dagcell{High Resolution Image} \\

II & \dagcell{High Resolution Image} & \dmedge & \dagcell{Geospatial Object Segmentation} \\

II & \dagcell{Geospatial Object Segmentation} & \mdedge & \dagcell{Segmentation Map} \\

II & \dagcell{Segmentation Map} & \dmedge & \dagcell{GeoChat} \\

II & \dagcell{High Resolution Image} & \dmedge & \dagcell{Multi-Spectral Classifier} \\

II & \dagcell{Multi-Spectral Classifier} & \mdedge & \dagcell{Image Classification Category} \\

II & \dagcell{Image Classification Category} & \dmedge & \dagcell{GeoChat} \\

II & \dagcell{High Resolution Image} & \dmedge & \dagcell{Anomaly Detection Forest} \\

II & \dagcell{Anomaly Detection Forest} & \mdedge & \dagcell{Anomaly Map Forest} \\

II & \dagcell{Anomaly Map Forest} & \dmedge & \dagcell{GeoChat} \\

II & \dagcell{High Resolution Image} & \dmedge & \dagcell{Landslide Segmentation} \\

II & \dagcell{Landslide Segmentation} & \mdedge & \dagcell{Landslide Map} \\

II & \dagcell{Landslide Map} & \dmedge & \dagcell{GeoChat} \\

II & \dagcell{High Resolution Image} & \dmedge & \dagcell{Urban Anomaly Detection} \\

II & \dagcell{Urban Anomaly Detection} & \mdedge & \dagcell{Anomaly Map Urban} \\

II & \dagcell{Anomaly Map Urban} & \dmedge & \dagcell{GeoChat} \\

II & \dagcell{GeoChat} & \mdedge & \dagcell{Contextual Description} \\

II & {[}\dagcell{Anomaly Map Urban} + \dagcell{Landslide Map}{]} & \dmedge & \dagcell{GeoChat} \\

II & {[}\dagcell{Landslide Map} + \dagcell{Anomaly Map Forest}{]} & \dmedge & \dagcell{GeoChat} \\

II & {[}\dagcell{Landslide Map} + \dagcell{Anomaly Map Urban}{]} & \dmedge & \dagcell{GeoChat} \\

II & {[}\dagcell{Landslide Map} + \dagcell{Anomaly Map Urban} + \dagcell{Anomaly Map Forest}{]} & \dmedge & \dagcell{GeoChat} \\

\bottomrule
\end{tabularx}

\caption{Typed edges in the DisasterBench DAG by subgraph (Part II).}

\end{table*}
}

{
\setlength{\tabcolsep}{3pt}
\renewcommand{\arraystretch}{0.98}
\scriptsize

\begin{table*}[t]
\centering
\begin{tabularx}{\textwidth}{@{}>{\centering\arraybackslash}p{0.045\textwidth}>{\raggedright\arraybackslash}X>{\centering\arraybackslash}p{0.07\textwidth}>{\raggedright\arraybackslash}X@{}}
\toprule
\textbf{G.} & \textbf{Source} & \textbf{Type} & \textbf{Target} \\
\midrule

\multicolumn{4}{@{}l}{\textit{Subgraph III: Adverse-Weather Perception (18 edges)}} \\

III & \dagcell{Metadata Prompt Image Generation} & \mdedge & \dagcell{High Resolution Image} \\

III & \dagcell{High Resolution Image} & \dmedge & \dagcell{Foggy Scenario Object Detection} \\

III & \dagcell{Foggy Scenario Object Detection} & \mdedge & \dagcell{Detected Objects Foggy} \\

III & \dagcell{Detected Objects Foggy} & \dmedge & \dagcell{GeoChat} \\

III & \dagcell{GeoChat} & \mdedge & \dagcell{Contextual Description} \\

III & \dagcell{High Resolution Image} & \dmedge & \dagcell{Low-Light Object Detection} \\

III & \dagcell{Low-Light Object Detection} & \mdedge & \dagcell{Detected Objects Low Light} \\

III & \dagcell{High Resolution Image} & \dmedge & \dagcell{Crowd Counting in Adverse Weather} \\

III & \dagcell{Crowd Counting in Adverse Weather} & \mdedge & \dagcell{Counted Crowd} \\

III & \dagcell{High Resolution Image} & \dmedge & \dagcell{Weather Degraded Image Restoration} \\

III & \dagcell{Weather Degraded Image Restoration} & \mdedge & \dagcell{Restored Image} \\

III & \dagcell{Restored Image} & \dmedge & \dagcell{Foggy Scenario Object Detection} \\

III & \dagcell{Foggy Scenario Object Detection} & \mdedge & \dagcell{Detected Objects Foggy} \\

III & \dagcell{Detected Objects Foggy} & \dmedge & \dagcell{GeoChat} \\

III & \dagcell{Restored Image} & \dmedge & \dagcell{Low-Light Object Detection} \\

III & \dagcell{Low-Light Object Detection} & \mdedge & \dagcell{Detected Objects Low Light} \\

III & \dagcell{Restored Image} & \dmedge & \dagcell{Crowd Counting in Adverse Weather} \\

III & \dagcell{Crowd Counting in Adverse Weather} & \mdedge & \dagcell{Counted Crowd} \\

\bottomrule
\end{tabularx}

\caption{Typed edges in the DisasterBench DAG by subgraph (Part III).}

\end{table*}
}

{
\setlength{\tabcolsep}{3pt}
\renewcommand{\arraystretch}{0.98}
\scriptsize

\begin{table*}[t]
\centering
\begin{tabularx}{\textwidth}{@{}>{\centering\arraybackslash}p{0.045\textwidth}>{\raggedright\arraybackslash}X>{\centering\arraybackslash}p{0.07\textwidth}>{\raggedright\arraybackslash}X@{}}
\toprule
\textbf{G.} & \textbf{Source} & \textbf{Type} & \textbf{Target} \\
\midrule

\multicolumn{4}{@{}l}{\textit{Subgraph IV: Hydrological Modeling (7 edges)}} \\

IV & \dagcell{Precipitation Nowcasting} & \mdedge & \dagcell{Predicted Precipitation} \\

IV & \dagcell{Predicted Precipitation} & \dmedge & \dagcell{Precipitation Data Conversion Tool} \\

IV & \dagcell{Precipitation Data Conversion Tool} & \mdedge & \dagcell{Converted Precipitation Data} \\

IV & \dagcell{Converted Precipitation Data} & \dmedge & \dagcell{Flood Depth Prediction} \\

IV & \dagcell{Flood Depth Prediction} & \mdedge & \dagcell{Predicted Water Depths} \\

IV & \dagcell{Predicted Water Depths} & \dmedge & \dagcell{Depth-Speed Model} \\

IV & \dagcell{Depth-Speed Model} & \mdedge & \dagcell{Traffic Speed} \\

\bottomrule
\end{tabularx}

\caption{Typed edges in the DisasterBench DAG by subgraph (Part IV).}

\end{table*}
}

\normalsize
\renewcommand{\arraystretch}{1.0}

\section{Dataset Statistics}
\label{app:dataset_stats}

Table~\ref{tab:task_stats} summarizes the structural composition of the
233 benchmark tasks. Tasks are categorized into three types following
the taxonomy of \citet{shen2024taskbench}: \textit{Node} (single-step),
\textit{Chain} (sequential multi-step), and \textit{Branching}
(one-to-many dependency). Compositional difficulty increases across
types: Node tasks test tool grounding in isolation, while Chain and
Branching tasks require parameter propagation and dependency reasoning
across multiple steps.

\begin{table}[h]
\centering
\footnotesize
\setlength{\tabcolsep}{3.5pt}
\begin{tabular*}{\columnwidth}{@{\extracolsep{\fill}}lcccc}
\toprule
\textbf{Type} & \textbf{Count} & \textbf{\% of Total}
& \textbf{Depth Range} & \textbf{Avg.\ Depth} \\
\midrule
Node      & 35  & 15.0\% & 1     & 1.00 \\
Chain     & 166 & 71.2\% & 2--9  & 2.79 \\
Branching & 32  & 13.7\% & 2--5  & 2.75 \\
\midrule
\textbf{Total} & 233 & 100\% & 1--9 & 2.53 \\
\bottomrule
\end{tabular*}
\caption{Structural composition of DisasterBench tasks.}
\label{tab:task_stats}
\end{table}

\textbf{Tool coverage.}
The 35 Node tasks span 21 distinct tools. To assess robustness to
surface-level variation independently of compositional reasoning, 12
tools are instantiated across 2--3 task variants with distinct
natural-language descriptions and input parameterizations.
Table~\ref{tab:agent_freq} (Appendix~\ref{app:coverage}) reports the
full per-tool invocation frequency across all 233 tasks.

\textbf{Plan depth distribution.}
Figure~\ref{fig:depth_accuracy} plots exact-match accuracy as a
function of plan depth. The majority of tasks fall at depths 2--4
($n = 167$), with three tasks at depths $\geq 5$ (depths 5, 6, and 9).
These deeper tasks are aggregated into a single bucket in the depth
analysis due to limited per-bucket sample size.

\textbf{Random baseline bound.}
As a sanity check on benchmark non-triviality, the probability of
matching a ground-truth tool sequence by uniform random selection from
$N=26$ tools over a plan of length $L$ is $N^{-L}$. At the
benchmark's average depth $\bar{L}=2.53$, this yields
$26^{-2.53} \approx 2.6 \times 10^{-4}$, confirming that correct
planning cannot be achieved by chance. All structural baselines
empirically achieve $0\%$ EM (Section~\ref{sec:experimental_setup}).

\subsection{Tool Semantic Overlap Analysis}
\label{app:agent_overlap}

To verify that benchmark difficulty arises from semantic ambiguity
among tools rather than from pool size alone, we compute pairwise
cosine similarity between tool descriptions using two sentence
encoders: \texttt{all-MiniLM-L6-v2} and \texttt{intfloat/e5-small-v2}.
tool descriptions are the natural-language functional specifications
provided to models during evaluation.

Table~\ref{tab:agent_overlap} reports similarity statistics across all
$\binom{26}{2} = 325$ tool pairs. The two encoders reveal
complementary perspectives: under MiniLM, $18.5\%$ of pairs exceed
cosine $0.50$ and only $0.6\%$ exceed $0.80$, reflecting a conservative
similarity distribution; under E5-small, all 325 pairs exceed $0.77$
and $92.9\%$ exceed $0.80$, indicating that tool descriptions are
uniformly close in the E5 embedding space. Both encoders agree that the
most semantically similar pairs are operationally distinct tools:
\texttt{Temporal\_Image\_Sequence\_Classifier} $\leftrightarrow$
\texttt{RGB\_GeoImage\_Classifier} (MiniLM: $0.855$; E5: $0.965$),
\texttt{Anomaly\_Detection\_Forest} $\leftrightarrow$
\texttt{Urban\_Anomaly\_Detection} (MiniLM: $0.804$; E5: $0.933$), and
\texttt{Foggy\_Scenario\_Object\_Detection} $\leftrightarrow$
\texttt{Low-Light\_Object\_Detection} (MiniLM: $0.783$; E5: $0.952$).
These pairs appear semantically interchangeable in natural language yet
differ substantially in operational input requirements and downstream
compatibility.

\begin{table}[h]
\centering
\footnotesize
\setlength{\tabcolsep}{3.5pt}
\begin{tabular*}{\columnwidth}{@{\extracolsep{\fill}}lcc}
\toprule
\textbf{Metric} & \textbf{MiniLM} & \textbf{E5-small} \\
\midrule
Mean cosine similarity   & 0.350 & 0.850 \\
Median cosine similarity & ---   & 0.847 \\
Std                      & ---   & 0.035 \\
Min / Max                & --- / 0.855 & 0.772 / 0.965 \\
\midrule
Pairs $\geq 0.50$ & 60 (18.5\%)  & 325 (100.0\%) \\
Pairs $\geq 0.70$ & 6 \ (1.8\%)  & 325 (100.0\%) \\
Pairs $\geq 0.80$ & 2 \ (0.6\%)  & 302 \ (92.9\%) \\
Pairs $\geq 0.90$ & ---          & 23 \ \ (7.1\%) \\
Pairs $\geq 0.95$ & ---          & 2 \ \ \ (0.6\%) \\
\bottomrule
\end{tabular*}
\caption{Pairwise cosine similarity statistics across all 325 agent
pairs under two sentence encoders. Both encoders confirm high semantic
overlap among operationally distinct tools, supporting the claim that
benchmark difficulty arises from fine-grained semantic disambiguation
rather than pool size.}
\label{tab:agent_overlap}
\end{table}

These results confirm that the 26-tool pool exhibits substantial
description-level semantic overlap despite operational divergence.
The uniformly high E5 similarities ($\geq 0.77$ for all pairs) further
suggest that surface-level semantic matching alone is insufficient for
correct tool grounding, models must reason about operational
compatibility and downstream dependency constraints beyond description
similarity.

\section{Tool Invocation Frequency}
\label{app:coverage}

To verify balanced task coverage, we report the number of tasks in which
each tool appears at least once in the ground-truth plan
(Table~\ref{tab:agent_freq}). No single tool dominates the distribution:
\texttt{GeoChat} is the most frequent at 54 tasks (23.2\%), while 18 of
the 26 tools each appear in fewer than 20\% of tasks, confirming that
the benchmark exercises the full tool pool.

\begin{table*}[t]
\centering
\footnotesize
\setlength{\tabcolsep}{3.5pt}
\begin{tabular*}{\textwidth}{@{\extracolsep{\fill}}p{0.72\textwidth}cc}
\toprule
\textbf{Tool} & \textbf{Tasks (\#)} & \textbf{Tasks (\%)} \\
\midrule
Urban\_Anomaly\_Detection & 55 & 23.6 \\
GeoChat & 54 & 23.2 \\
Anomaly\_Detection\_Forest & 49 & 21.0 \\
Geospatial\_Object\_Segmentation & 47 & 20.2 \\
Landslide\_Segmentation & 39 & 16.7 \\
Crowd\_Counting\_in\_Adverse\_Weather & 38 & 16.3 \\
Weather\_Degraded\_Image\_Restoration & 32 & 13.7 \\
Low-Light\_Object\_Detection & 29 & 12.4 \\
Foggy\_Scenario\_Object\_Detection & 28 & 12.0 \\
Building\_damage\_assessment & 27 & 11.6 \\
Change\_Mapping\_and\_Detection & 24 & 10.3 \\
Multi\_Spectral\_Classifier & 19 & 8.2 \\
High-Resolution\_Image\_Reconstructor & 16 & 6.9 \\
Metadata\_and\_Text\_Prompt\_Image\_Generation & 15 & 6.4 \\
Temporal\_High\_Resolution\_Image\_Generation & 14 & 6.0 \\
Temporal\_Image\_Sequence\_Classifier & 12 & 5.2 \\
Flood\_depth\_prediction & 10 & 4.3 \\
RGB\_GeoImage\_Classifier & 9 & 3.9 \\
Precipitation\_Nowcasting & 7 & 3.0 \\
precipitation\_data\_convert\_tool & 7 & 3.0 \\
depth\_speed\_model & 3 & 1.3 \\
Event\_detection & 2 & 0.9 \\
Multimodal\_mobility\_prediction\_under\_events & 2 & 0.9 \\
Post\_Disaster\_Mobility\_Recovery & 2 & 0.9 \\
Toponym\_Detection & 2 & 0.9 \\
Video\_anomaly\_detection & 2 & 0.9 \\
\bottomrule
\end{tabular*}
\caption{Tool Invocation Frequency across 233 benchmark tasks.}
\label{tab:agent_freq}
\end{table*}

\section{Evaluation Details}
\label{app:eval}

This appendix provides the formal definitions of the evaluation metrics
introduced in Section~\ref{sec:eval_protocol}, including the partial-credit
diagnostic metrics, the First-Point-of-Failure (FPoF) error taxonomy, and
the implication structure among metrics.

\subsection{Partial-Credit Diagnostic Metrics}
\label{app:partial_credit_def}

All four metrics are task-level booleans: each returns $1$ if the
corresponding condition holds for every step in the predicted plan, and $0$
otherwise. Dataset-level scores are obtained by averaging over all 233
tasks (i.e., the fraction of tasks for which the condition holds).

Let $\hat{P} = (\hat{s}_1, \dots, \hat{s}_{\hat{T}})$ denote the predicted
plan and $P = (s_1, \dots, s_T)$ the ground-truth plan. We first require
$\hat{T} = T$ (plan lengths match); if not, all four metrics return $0$.
When $\hat{T} = T$, each step $s_t$ is a tuple
$(\texttt{step}_t, \texttt{agent}_t, \texttt{inputs}_t,
\texttt{outputs}_t, \texttt{dep}_t, \texttt{dep\_content}_t)$, and the
metrics are defined as follows.

\paragraph{Tool Accuracy (A).}
All step indices and agent identities match:
\[
\resizebox{\columnwidth}{!}{$\displaystyle
\text{A}(\hat{P}, P) = \mathbb{I}\!\left(\forall\, t:\, \hat{\texttt{step}}_t = \texttt{step}_t \;\land\; \hat{\texttt{agent}}_t = \texttt{agent}_t\right)
$}
\]

\paragraph{Parameter Accuracy.}
tool conditions hold, and inputs and outputs match at every step:
\[
\resizebox{\columnwidth}{!}{$\displaystyle
\text{Param}(\hat{P}, P) = \mathbb{I}\!\left(\text{A} = 1 \;\land\; \forall\, t:\, \hat{\texttt{inputs}}_t = \texttt{inputs}_t \;\land\; \hat{\texttt{outputs}}_t = \texttt{outputs}_t\right)
$}
\]

\paragraph{Dependency Accuracy.}
tool conditions hold, and dependency structure and dependency content
match at every step:
\[
\resizebox{\columnwidth}{!}{$\displaystyle
\text{Dep}(\hat{P}, P) = \mathbb{I}\!\left(\text{A} = 1 \;\land\; \forall\, t:\, \text{sort}(\hat{\texttt{dep}}_t) = \text{sort}(\texttt{dep}_t) \;\land\; \hat{\texttt{dep\_content}}_t \doteq \texttt{dep\_content}_t\right)
$}
\]
where $\text{sort}(\cdot)$ normalizes dependency lists to sorted order
(allowing $-1$ vs.\ $[-1]$ equivalence), and $\doteq$ denotes deep
equality with both-null treated as equal.

\paragraph{Exact Match (EM).}
EM is computed by exact structural match over all step fields under the
canonical step schema; equivalently, $\text{EM} = 1$ if and only if both
Parameter Accuracy and Dependency Accuracy equal $1$:
\[
\text{EM}(\hat{P}, P) = \mathbb{I}\!\left(
\text{Param} = 1 \;\land\; \text{Dep} = 1
\right)
\]

\subsection{Semantic-Execution Gap: Full Results}
\label{app:semantic_gap}

Table~\ref{tab:partial_credit} reports Tool Accuracy (A) and
exact-match accuracy (EM) for all 14 models across five planning
methods. The gap between A and EM---the \textbf{semantic-execution
gap}---is present for every model under every method, confirming that
semantically plausible tool selection is consistently easier than
execution-consistent workflow generation. These results support the
analysis in Section~\ref{sec:gap}; notable cases discussed in the
main text include Llama-3.3-70B under DP ($22.31$pp gap) and
DeepSeek-V3.2 under CoT, where both A and EM drop uniformly relative
to its DP baseline.

\begin{table*}[t]
\centering
\small
\setlength{\tabcolsep}{2.5pt}
\renewcommand{\arraystretch}{1.0}
\begin{tabular*}{\textwidth}{@{\extracolsep{\fill}}llcccccccccc}
\toprule
& & \multicolumn{2}{c}{\textbf{DP}}
  & \multicolumn{2}{c}{\textbf{CoT}}
  & \multicolumn{2}{c}{\textbf{ToT}}
  & \multicolumn{2}{c}{\textbf{RAP}}
  & \multicolumn{2}{c}{\textbf{ReAct}} \\
\cmidrule(lr){3-4}\cmidrule(lr){5-6}\cmidrule(lr){7-8}\cmidrule(lr){9-10}\cmidrule(lr){11-12}
\textbf{Category} & \textbf{Model}
  & A & EM & A & EM & A & EM & A & EM & A & EM \\
\midrule
\multirow{2}{*}{Frontier}
& Gemini 3.1 Pro Preview
  & 75.97 & 68.24  & 81.97 & 72.53  & 82.40 & \textbf{73.39}  & 82.40 & 72.10  & 79.83 & 69.96 \\
& GPT-5.4
  & 60.09 & 54.08  & 69.96 & 62.23  & 69.96 & 62.23  & 77.25 & \textbf{65.67}  & 70.82 & 64.38 \\
\midrule
\multirow{7}{*}{\shortstack{Strong\\Open-source}}
& DeepSeek-V3.2
  & 68.24 & 57.94  & 41.20 & 36.05  & 56.65 & 50.21  & 40.77 & 34.33  & 67.81 & \textbf{61.80} \\
& DeepSeek-R1
  & 60.52 & 54.08  & 69.10 & \textbf{62.66}  & 61.80 & 53.22  & 38.63 & 35.19  & 60.52 & 56.22 \\
& Qwen3-Max
  & 62.66 & 55.79  & 71.67 & 62.66  & 73.39 & \textbf{63.95}  & 73.39 & 63.09  & 67.38 & 62.23 \\
& Qwen3-Max-Thinking
  & 64.38 & 57.51  & 73.82 & \textbf{66.52}  & 68.67 & 62.23  & 71.24 & 61.37  & 68.24 & 61.37 \\
& Qwen3.5-27B
  & 62.23 & 58.37  & 67.38 & \textbf{62.23}  & 66.09 & 59.23  & 66.09 & 58.80  & 64.38 & 59.23 \\
& Gemma-4-31B
  & 73.82 & \textbf{66.52}  & 69.96 & 63.09  & 65.67 & 58.37  & 60.09 & 50.64  & 68.24 & 62.23 \\
& Llama-3.3-70B
  & 60.94 & 38.63  & 66.95 & \textbf{58.80}  & 51.50 & 41.63  & 31.76 & 28.76  & 51.93 & 38.63 \\
\midrule
\multirow{5}{*}{\shortstack{Efficiency-\\oriented}}
& Qwen3.5-9B
  & 53.65 & 45.92  & 67.38 & 61.37  & 66.52 & 59.23  & 47.21 & 42.92  & 74.25 & \textbf{65.67} \\
& Llama-3.1-8B
  & 24.03 & 15.88  & 32.62 & 26.18  & 29.18 & 16.31  & 42.06 & \textbf{28.33}  & 31.76 & 26.61 \\
& Ministral-14B
  & 30.90 & 20.17  & 38.63 & 33.48  & 24.03 & 16.31  & 24.89 & 18.88  & 56.22 & \textbf{50.64} \\
& Ministral-8B
  & 28.76 & 25.32  & 34.76 & 31.33  & 27.04 & 21.03  & 18.88 & 12.88  & 51.07 & \textbf{44.21} \\
& Ministral-3B
  & 16.74 & 10.30  & 27.90 & 21.03  & 39.06 & 18.88  & 13.30 &  3.86  & 42.49 & \textbf{36.91} \\
\bottomrule
\end{tabular*}
\caption{Tool Accuracy (A) and exact-match accuracy (EM, \%) on
DisasterBench across all methods. The gap between A and EM reflects
the \textbf{semantic-execution gap}: failures that occur after correct
tool selection due to parameter binding or dependency propagation
errors. Both metrics are task-level booleans averaged over all 233
tasks. \textbf{Bold} indicates the best EM per model.}
\label{tab:partial_credit}
\end{table*}

\subsection{Metric Implication Structure}
\label{app:metric_implications}

The four metrics form a partial order under logical implication. An arrow
$X \Rightarrow Y$ means that $X = 1$ logically guarantees $Y = 1$ for
every task.

\[
\resizebox{0.88\columnwidth}{!}{$\displaystyle
\begin{array}{c}
\boxed{\text{EM}} \\
\swarrow \quad \searrow \\
\boxed{\text{Parameter Accuracy}} \qquad
\boxed{\text{Dependency Accuracy}} \\
\searrow \quad \swarrow \\
\boxed{\text{Tool Accuracy}}
\end{array}$}
\]

\noindent
Formally:
\begin{itemize}
    \item $\text{Param} = 1 \;\Rightarrow\; \text{A} = 1$
          \quad (parameter check subsumes step and tool matching).
    \item $\text{Dep} = 1 \;\Rightarrow\; \text{A} = 1$
          \quad (dependency check subsumes step and tool matching).
    \item $\text{EM} = 1 \;\Rightarrow\; \text{Param} = 1 \;\land\;
          \text{Dep} = 1 \;\land\; \text{A} = 1$.
\end{itemize}

\noindent
Crucially, Parameter Accuracy and Dependency Accuracy are \emph{not}
mutually comparable: a task can satisfy one without the other, because they
check disjoint subsets of step attributes (inputs/outputs vs.\
dependency structure/content). This diamond structure enables independent
diagnosis of parameter binding failures and dependency reasoning failures
on the same dataset.

\subsection{FPoF Error Taxonomy}
\label{app:fpof_taxonomy}

The First-Point-of-Failure (FPoF) analyzer examines each failed prediction
and identifies the earliest step at which the predicted plan diverges from
the ground truth. The error at this step is classified into one of eight
fine-grained types (left column of Table~\ref{tab:fpof_mapping}), which we
aggregate into three top-level categories for the analysis in
Section~\ref{sec:failure_modes}.

\begin{table*}[t]
\centering
\small
\setlength{\tabcolsep}{3pt}
\begin{tabular*}{\textwidth}{@{\extracolsep{\fill}}p{0.38\textwidth}
p{0.28\textwidth}p{0.28\textwidth}}
\toprule
\textbf{Evaluator Error Type} & \textbf{Top-Level Category}
& \textbf{Condition} \\
\midrule
\texttt{agent\_mismatch}
  & tool mismatch
  & Predicted tool $\neq$ gold tool \\
\midrule
\texttt{parameter\_error}
  & Parameter Binding Error
  & Agent matches but inputs or outputs differ \\
\texttt{dependency\_error}
  & Parameter Binding Error
  & Agent matches but dependency list differs \\
\texttt{dependency\_content\_error}
  & Parameter Binding Error
  & Agent matches but dependency content differs \\
\midrule
\texttt{hallucinated\_extra\_steps}
  & Structural Error
  & Predicted plan is longer than gold
    (and all aligned steps match) \\
\texttt{early\_stop}
  & Structural Error
  & Predicted plan is shorter than gold
    (and all aligned steps match) \\
\texttt{empty\_output}
  & Structural Error
  & Model produces no parseable plan \\
\texttt{format\_error}
  & Structural Error
  & Output is parseable but violates JSON schema \\
\bottomrule
\end{tabular*}
\caption{Mapping from fine-grained evaluator error types to the three
top-level FPoF categories used in the main text.}
\label{tab:fpof_mapping}
\end{table*}

\paragraph{Classification procedure.}
For a failed prediction $\hat{P}$ with ground truth $P$:
\begin{enumerate}
    \item \textbf{Structural pre-check.} If $\hat{P}$ is empty (no model
          output), return \texttt{empty\_output}. If $\hat{P}$ cannot be
          parsed as a JSON list, return \texttt{format\_error}.

    \item \textbf{Step-level scan.} Let
          $L = \min(\hat{T}, T)$. For each aligned step
          $t = 0, \dots, L{-}1$, check in order:
    \begin{enumerate}
        \item Step index mismatch
              ($\hat{\texttt{step}}_t \neq \texttt{step}_t$):
              return \texttt{parameter\_error} at step $t$.
        \item tool mismatch
              ($\hat{\texttt{agent}}_t \neq \texttt{agent}_t$):
              return \texttt{agent\_mismatch} at step $t$.
        \item Inputs or outputs mismatch: return
              \texttt{parameter\_error} at step $t$.
        \item Dependency list mismatch (after sorting): return
              \texttt{dependency\_error} at step $t$.
        \item Dependency content mismatch: return
              \texttt{dependency\_content\_error} at step $t$.
    \end{enumerate}

    \item \textbf{Length mismatch (only if all aligned steps match).}
          If $\hat{T} < T$, return \texttt{early\_stop}. If
          $\hat{T} > T$, return \texttt{hallucinated\_extra\_steps}.
          Otherwise ($\hat{T} = T$ and all steps match), the prediction
          is correct.
\end{enumerate}

\noindent
This priority order ensures that more fundamental errors (wrong agent) are
reported before finer-grained ones (wrong parameters, wrong dependencies).
Length mismatch is attributed only when all overlapping steps are
correct---capturing plans that are structurally valid up to the point of
premature termination or over-extension. By reporting only the \emph{first} point of failure, FPoF isolates 
root causes from cascading downstream errors that typically follow 
once a plan has diverged from the ground-truth workflow.

\paragraph{Codebase correspondence.}
The fine-grained error types correspond directly to the return values of
\texttt{Evaluator.analyze\_error\_propagation()} in
\texttt{evaluators/evaluators.py}. The field \texttt{tools\_correct} in
the released prediction files corresponds to Tool Accuracy as defined
above.

\section{Datasheet for DisasterBench}
\label{app:datasheet}

Following the datasheet framework of~\citep{gebru2021datasheets},
we provide the following documentation for DisasterBench.

\paragraph{Motivation.}
DisasterBench was created to evaluate the ability of LLMs to coordinate
multi-step agent workflows in disaster management, a setting where
incorrect parameter bindings or dependency propagation can invalidate
downstream workflow execution. The benchmark is intended for academic
research on workflow grounding and structured multi-agent planning.

\paragraph{Composition.}
The dataset consists of 233 planning tasks, each comprising a
natural-language task description and a structured ground-truth plan in
JSON format specifying tool calls, input parameters, dependency
structure, and dependency content. Tasks are derived from four
disaster-response workflow categories: temporal remote sensing analysis,
image reconstruction, adverse-weather perception, and hydrological
modeling. No personally identifiable information is included.

\paragraph{Collection process.}
Tasks were generated through a two-stage pipeline. First, GPT-4o
(\texttt{gpt-4o}, accessed January--March 2025) was used to generate
naturalistic task descriptions from sampled workflow structures.
Second, all generated tasks were manually verified through two rounds of
expert review by annotators with domain familiarity in remote sensing
and disaster management. The workflow dependency structure was
expert-curated to ensure semantic and technical validity of all agent
dependencies.

\paragraph{Preprocessing / cleaning / labeling.}
Ground-truth workflows were verified for semantic consistency between
task descriptions and target plans, as well as strict adherence to agent
input/output schemas. Ambiguous cases were resolved through group
discussion, and unresolved tasks were discarded.

\paragraph{Uses.}
DisasterBench is intended for evaluating workflow grounding and
structured planning capabilities of LLMs under executable dependency
constraints. The benchmark should not be used for training without
careful consideration of potential benchmark contamination. The
disaster-management setting is intended as a research testbed and does
not constitute operational guidance for real-world disaster response.

\paragraph{Distribution.}
The dataset and evaluation code will be publicly released upon
acceptance under an open-source license.

\paragraph{Maintenance.}
The dataset will be maintained by the authors. Issues, corrections, and
future updates will be managed through the public project repository.

\paragraph{Known limitations.}
Most tasks in DisasterBench are linear workflows, which may
under-represent highly parallel or data-merging coordination patterns.
Evaluation is plan-level only and does not include execution-grounded
runtime verification. In addition, the tool pool reflects a curated
subset of disaster-response capabilities and may not cover all real-world
workflow configurations.

\section{Prompt Templates}
\label{app:prompts}

This appendix provides the prompt templates used throughout the benchmark
construction and evaluation pipeline. Section~\ref{app:prompt_taskgen}
covers the task generation prompt used during benchmark construction
(Section~\ref{sec:task_generation}), Section~\ref{app:filter_prompt}
covers the agent filtering prompt (Section~\ref{sec:agent_pool}), and
Sections~\ref{app:prompt_schema}--\ref{app:prompt_react} cover the five
planning method prompts used during evaluation
(Section~\ref{sec:experimental_setup}). All planning method prompts share
the same agent description block (\texttt{\{agents\_desc\}}), which
contains the full input/output schema for all 26 tools. The workflow
structure is \emph{not} included in any prompt; models must infer
valid compositions from agent specifications alone. For brevity, the
shared JSON output schema is shown once
(Section~\ref{app:prompt_schema}) and omitted from subsequent templates.
Complete prompt texts including all few-shot examples are provided in the
released codebase.

\subsection{Task Description Generation (Self-Instruct)}
\label{app:prompt_taskgen}

During benchmark construction (Section~\ref{sec:task_generation}), we use
GPT-4o to generate naturalistic task descriptions from sampled ground-truth
plans via the following prompt:

\begin{lstlisting}[basicstyle=\ttfamily\scriptsize, breaklines=true,
  frame=single]
You are an expert technical writer specializing in disaster
management scenarios. You will be given a structured agent
execution plan (in JSON format) that describes a sequence
of AI agents, their inputs, outputs, and inter-step
dependencies for a disaster-response workflow.

Your task is to write a realistic, goal-oriented task
description that a disaster management practitioner might
submit to an AI planning system. The description should:

1. Describe the user's high-level goal and the disaster
   scenario context (e.g., flood assessment, post-earthquake
   damage analysis, storm impact evaluation).
2. Mention the specific input data files that the user
   would provide (using the file paths from the plan's
   first step inputs).
3. Be written in natural language as a single coherent
   paragraph.

CRITICAL CONSTRAINTS:
- Do NOT mention any tool names, model names, or tool
  identifiers.
- Do NOT expose the number of steps, the plan structure,
  or the sequential ordering of operations.
- Do NOT reference intermediate data types, output keys,
  or dependency relationships.
- The description should read as if written by a domain
  practitioner who knows what outcome they want but not
  how to achieve it technically.

Input plan:
{ground_truth_plan_json}

Write the task description below:
\end{lstlisting}

\noindent
All generated descriptions undergo two rounds of expert review to ensure
that no structural cues are leaked, as described in
Section~\ref{sec:task_generation}.

\subsection{Agent Filtering}
\label{app:filter_prompt}

During tool pool construction (Section~\ref{sec:agent_pool}), candidate
models are screened for inclusion using the following prompt:

\begin{lstlisting}[basicstyle=\ttfamily\scriptsize, breaklines=true,
  frame=single]
You are an expert in AI for disaster management and remote
sensing. You will be given information about a publicly
available AI model (title, abstract, and repository link).
Your task is to determine whether this model is suitable
for inclusion in a disaster-response agent benchmark.

A model is suitable if ALL of the following criteria are
met:

1. DOMAIN RELEVANCE: The model addresses a task directly
   applicable to disaster response workflows, such as:
   remote sensing image analysis, damage assessment,
   flood/precipitation modeling, anomaly detection, change
   detection, crowd monitoring, object detection under
   adverse conditions, geospatial reasoning, or disaster-
   related text/event processing.

2. TYPED INTERFACE: The model has clearly defined input and
   output types (e.g., takes a satellite image and outputs
   a segmentation map), enabling it to be composed with
   other models in a typed pipeline.

3. IMPLEMENTATION AVAILABILITY: The model has a publicly
   accessible codebase or pre-trained weights that can be
   used for inference.

For each candidate, respond with:
- Decision: INCLUDE or EXCLUDE
- Justification: One sentence explaining why.

Candidate model:
Title: {paper_title}
Abstract: {paper_abstract}
Repository: {repo_url}
\end{lstlisting}

\noindent
Candidates passing the automated screen are further verified manually for
implementation quality and interface compatibility, as described in
Section~\ref{sec:agent_pool}.

\subsection{Shared Output Schema (Planning Methods)}
\label{app:prompt_schema}

All five planning methods instruct the model to produce a JSON array
conforming to the following schema. Each element represents one planning
step:

\begin{lstlisting}[basicstyle=\ttfamily\scriptsize, breaklines=true,
  frame=single]
{
  "type": "array",
  "items": {
    "type": "object",
    "properties": {
      "agent":   {"type": "string"},
      "step":    {"type": "integer"},
      "dependence": {
        "type": "array",
        "items": {"type": "integer"}
      },
      "dependence_content": {
        "oneOf": [{"type": "null"},
                  {"type": "object"}]
      },
      "inputs":  {"type": "object"},
      "outputs": {"type": "array",
                  "items": {"type": "string"}}
    },
    "required": ["agent", "step", "dependence",
      "dependence_content", "inputs", "outputs"]
  }
}
\end{lstlisting}

\noindent
Key conventions shared across all methods:
\begin{itemize}
    \item \texttt{dependence}: \texttt{[-1]} if no dependency on prior
          steps; otherwise a single step index (e.g., \texttt{[0]}).
    \item \texttt{dependence\_content}: maps the depended step index to
          a list of output keys consumed, or \texttt{null} if
          independent.
    \item Generated inputs use the placeholder format
          \texttt{<GENERATED>-\{step\}-<\{output\_key\}>}.
\end{itemize}

\subsection{Direct Prompting (DP)}
\label{app:prompt_dp}

DP uses no intermediate reasoning. The system prompt instructs the model
to output \emph{only} the structured plan with no explanation.

\paragraph{System prompt.}
The system prompt specifies the output format and agent usage rules,
and appends the full agent description block:
\begin{lstlisting}[basicstyle=\ttfamily\scriptsize, breaklines=true,
  frame=single]
You are a task planning expert agent specializing in
disaster response automation. Your goal is to transform a
high-level disaster management task into a structured JSON
plan. You MUST output ONLY the final structured plan, with
no reasoning and no extra text.

STRICT OUTPUT FORMAT (must follow):
- Output exactly one line that starts with:
  The structured task plan is:
- After that prefix, output a single valid JSON ARRAY only.
- Use double quotes for JSON keys and string values.
- Do not output any other text before or after the JSON.

IMPORTANT RULES:
- Choose the minimum set of agents required.
- Each step index is 0-based and sequential.
- Each step may depend on at most ONE previous step.
- Use dependence [-1] when all inputs come from the user.
- For generated inputs, use:
  "<GENERATED>-<STEP>-<OutputKey>"

Agents and Input Output Details:
{agents_desc}

[Shared output schema inserted here]
\end{lstlisting}

\paragraph{User prompt.}
The user prompt provides the task description and elicits the plan
directly without intermediate reasoning:
\begin{lstlisting}[basicstyle=\ttfamily\scriptsize, breaklines=true,
  frame=single]
Instruction: {task_desc}

Response:
\end{lstlisting}

\noindent
No few-shot examples are provided for DP.

\subsection{Chain-of-Thought (CoT)}
\label{app:prompt_cot}

CoT elicits intermediate reasoning before plan generation.

\paragraph{System prompt.}
The system prompt is identical to DP except: (1) the no-reasoning
instruction is replaced with \emph{``Add `The structured task plan is:'
then followed by a structured JSON at the end of your response''}, and
(2) two few-shot examples are appended demonstrating step-by-step
reasoning followed by a valid plan.

\paragraph{User prompt.}
The user prompt appends a CoT trigger suffix to elicit step-by-step
reasoning before plan generation:
\begin{lstlisting}[basicstyle=\ttfamily\scriptsize, breaklines=true,
  frame=single]
Instruction: {task_desc}

Response: Let's think step by step.
\end{lstlisting}

\noindent
The suffix \texttt{Let's think step by step.} serves as the CoT trigger.

\paragraph{Few-shot examples.}
Two examples are provided: (1) a two-step chain
(\texttt{Urban\_Anomaly\_Detection} $\rightarrow$ \texttt{GeoChat}) and
(2) a four-step branching plan
(\texttt{Metadata\_and\_Text\_Prompt\_Image\_Generation} $\rightarrow$
\texttt{Weather\_Degraded\_Image\_Restoration} $\rightarrow$
\texttt{Crowd\_Counting} $+$
\texttt{Low-Light\_Object\_Detection}). Each example includes a
complete reasoning trace followed by the structured JSON output.

\subsection{Tree-of-Thought (ToT)}
\label{app:prompt_tot}

ToT uses two separate prompts: a \emph{generation prompt} for the actor
LLM that proposes candidate plans, and a \emph{vote prompt} for the
critic LLM that selects the best candidate.

\paragraph{Generation prompt (actor).}
The generation prompt follows the same structure as CoT (system prompt
with tool descriptions, shared output schema, and two few-shot
examples). Multiple candidate plans are sampled at
\texttt{temperature}$=0.5$ (see Appendix~\ref{app:implementation} for
the branching factor and search depth configuration).

\paragraph{Vote prompt (critic).}
The critic receives all candidate plans and selects the best one:

\begin{lstlisting}[basicstyle=\ttfamily\scriptsize, breaklines=true,
  frame=single]
You are a task planning critic agent specializing in
rating the plans generated by an actor agent. Your task
is to choose the best planning based on the provided
choices.

You must check the following:
[Shared output schema and field definitions inserted here]

At the end, reply with exactly one line in this exact
format (no other text before or after):
The best choice is: <BEST_OPTION>
where <BEST_OPTION> is an integer from 1 to the number
of choices listed above.
\end{lstlisting}

\subsection{Reasoning via Planning (RAP)}
\label{app:prompt_rap}

RAP uses two prompt types within the MCTS loop: a
\emph{sub-question/sub-answer} prompt for node expansion, and a
\emph{usefulness} prompt for reward estimation.

\paragraph{Sub-question / sub-answer prompt.}
This prompt instructs the model to decompose the task into
sub-questions, answering each with a partial plan. When sufficient
context has been accumulated, the model begins the final sub-question
with \texttt{Now we can answer the question:} followed by the
consolidated plan.

\begin{lstlisting}[basicstyle=\ttfamily\scriptsize, breaklines=true,
  frame=single]
You are an expert disaster management AI agent tasked
with breaking down disaster management queries into
simpler, actionable questions. You will build the answer
in multiple steps, where in each step, you generate a
question based on the user question and previous answers.
When all required subquestions are answered, start the
next question with "Now we can answer the question:"
followed by the question.

[Shared output schema inserted here]

Agents description:
{agents_list}

[Three few-shot examples demonstrating sub-question
decomposition, each showing 2-3 sub-questions leading
to a final consolidated plan]
\end{lstlisting}

\paragraph{Usefulness prompt.}
The usefulness prompt evaluates whether a candidate sub-question is
relevant to the original task. It receives the original question, prior
sub-questions, and the candidate, then outputs \texttt{Yes} or
\texttt{No} with a one-sentence justification. Three few-shot examples
are provided demonstrating both useful and non-useful sub-questions.

\subsection{ReAct}
\label{app:prompt_react}

ReAct interleaves reasoning traces (\texttt{Thought}), action decisions
(\texttt{Action}), and state observations (\texttt{Observation}) before
producing the final plan.

\paragraph{System prompt.}
The system prompt specifies the Thought/Action/Observation reasoning
format and appends the full agent description block:
\begin{lstlisting}[basicstyle=\ttfamily\scriptsize, breaklines=true,
  frame=single]
You are a task planning expert agent specializing in
disaster response automation.

You MUST reason using the following format:

Thought: [analyze what tools are needed and why]
Action: [decide the next step: which agent to select]
Observation: [what this step produces and what the next
             step needs]
... (repeat until all steps are planned)

Then output the final structured plan on ONE line:
The structured task plan is: [JSON array]

[Shared output schema and rules inserted here]

Agents and Input Output Details:
{agents_desc}
\end{lstlisting}

\paragraph{User prompt.}
The user prompt instructs the model to apply the reasoning format
before producing the final structured plan:
\begin{lstlisting}[basicstyle=\ttfamily\scriptsize, breaklines=true,
  frame=single]
Instruction: {task_desc}

Use Thought/Action/Observation to reason through the plan
step by step, then output the final structured task plan.
Response:
\end{lstlisting}

\paragraph{Few-shot examples.}
Four examples are provided covering: (1) a two-step chain (image
restoration $\rightarrow$ object detection), (2) a three-step chain
(precipitation nowcasting $\rightarrow$ data conversion $\rightarrow$
flood prediction), (3) a two-step chain (anomaly detection
$\rightarrow$ contextual description), and (4) a five-step branching
plan with fan-out (image generation $\rightarrow$ parallel forest and
urban anomaly detection $\rightarrow$ two separate GeoChat
descriptions). Each example demonstrates the full
Thought/Action/Observation trace followed by the structured JSON output.

\section{Per-Model FPoF Breakdowns}
\label{app:fpof}

Table~\ref{tab:fpof_permodel} reports the First-Point-of-Failure (FPoF)
distribution for each of the 14 models under all five planning methods.
For each model--method combination, we show the number of perfectly
matched tasks (out of 233) and the top three most frequent FPoF error
types with counts and percentages of total tasks. Error types follow the
fine-grained taxonomy defined in Appendix~\ref{app:fpof_taxonomy}.

\begin{figure}[t]
    \centering
    \IfFileExists{fpof_distribution.pdf}{%
        \includegraphics[width=\columnwidth]{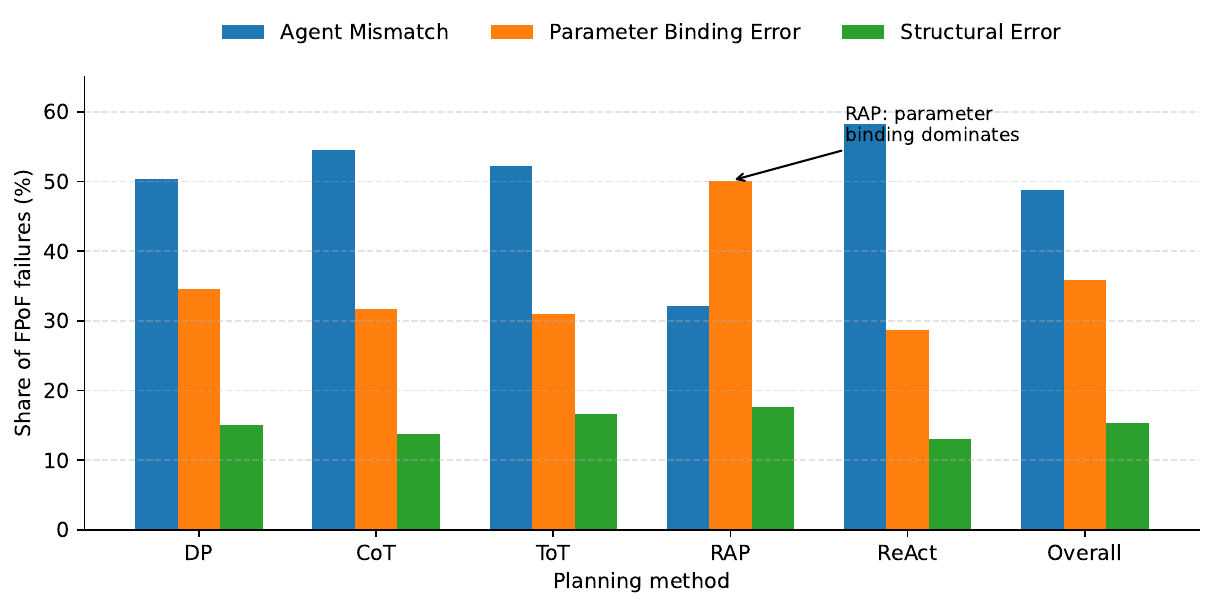}%
    }{%
        \fbox{\parbox[c][0.28\textheight][c]{0.95\columnwidth}{\centering
        FPoF figure placeholder\\(add
        \texttt{fpof_distribution.pdf})}}%
    }
    \caption{Distribution of First-Point-of-Failure (FPoF) error types
    across all 14 models and 5 planning methods, aggregated into the
    three top-level categories defined in
    Section~\ref{sec:eval_protocol}: tool mismatch ($48.8\%$),
    Parameter Binding Error ($35.8\%$), and Structural Error ($15.4\%$).
    RAP's distinctive early-termination signature ($11.1\%$ of RAP
    failures) is visible as an elevated Structural Error share.}
    \label{fig:fpof}
\end{figure}

{
\setlength{\tabcolsep}{2pt}
\renewcommand{\arraystretch}{0.92}
\scriptsize
\begin{table*}[t]
\centering
\begin{tabularx}{\textwidth}{@{}>{\raggedright\arraybackslash}p{0.17\textwidth}>{\centering\arraybackslash}p{0.085\textwidth}>{\centering\arraybackslash}p{0.085\textwidth}>{\raggedright\arraybackslash}X@{}}
\toprule
\textbf{Model} & \textbf{Method} & \textbf{Perfect} & \textbf{Top-3 FPoF Error Types} \\
\midrule
Gemini 3.1 Pro Preview   & DP & 159 & agent\_mismatch: 33 (14.2\%), parameter\_error: 32 (13.7\%), early\_stop: 7 (3.0\%) \\
    & CoT & 169 & agent\_mismatch: 30 (12.9\%), parameter\_error: 30 (12.9\%), halluc.\ steps: 3 (1.3\%) \\
    & ToT & 171 & agent\_mismatch: 34 (14.6\%), parameter\_error: 26 (11.2\%), halluc.\ steps: 2 (0.9\%) \\
    & RAP & 168 & parameter\_error: 33 (14.2\%), agent\_mismatch: 27 (11.6\%), early\_stop: 3 (1.3\%) \\
    & ReAct & 163 & agent\_mismatch: 39 (16.7\%), parameter\_error: 30 (12.9\%), halluc.\ steps: 1 (0.4\%) \\
\midrule
GPT-5.4   & DP & 126 & agent\_mismatch: 59 (25.3\%), parameter\_error: 31 (13.3\%), halluc.\ steps: 17 (7.3\%) \\
    & CoT & 145 & agent\_mismatch: 53 (22.7\%), parameter\_error: 31 (13.3\%), halluc.\ steps: 4 (1.7\%) \\
    & ToT & 145 & agent\_mismatch: 51 (21.9\%), parameter\_error: 30 (12.9\%), halluc.\ steps: 5 (2.1\%) \\
    & RAP & 153 & parameter\_error: 38 (16.3\%), agent\_mismatch: 29 (12.4\%), early\_stop: 9 (3.9\%) \\
    & ReAct & 150 & agent\_mismatch: 43 (18.5\%), parameter\_error: 33 (14.2\%), halluc.\ steps: 6 (2.6\%) \\
\midrule
DeepSeek-V3.2   & DP & 135 & agent\_mismatch: 55 (23.6\%), parameter\_error: 32 (13.7\%), halluc.\ steps: 10 (4.3\%) \\
    & CoT & 84 & parameter\_error: 114 (48.9\%), agent\_mismatch: 30 (12.9\%), halluc.\ steps: 4 (1.7\%) \\
    & ToT & 117 & agent\_mismatch: 73 (31.3\%), parameter\_error: 24 (10.3\%), halluc.\ steps: 17 (7.3\%) \\
    & RAP & 80 & early\_stop: 70 (30.0\%), parameter\_error: 53 (22.7\%), agent\_mismatch: 22 (9.4\%) \\
    & ReAct & 144 & agent\_mismatch: 59 (25.3\%), parameter\_error: 24 (10.3\%), early\_stop: 3 (1.3\%) \\
\midrule
DeepSeek-R1   & DP & 126 & agent\_mismatch: 59 (25.3\%), parameter\_error: 37 (15.9\%), early\_stop: 7 (3.0\%) \\
    & CoT & 146 & agent\_mismatch: 57 (24.5\%), parameter\_error: 24 (10.3\%), halluc.\ steps: 4 (1.7\%) \\
    & ToT & 124 & agent\_mismatch: 71 (30.5\%), parameter\_error: 29 (12.4\%), halluc.\ steps: 9 (3.9\%) \\
    & RAP & 82 & parameter\_error: 104 (44.6\%), agent\_mismatch: 32 (13.7\%), early\_stop: 13 (5.6\%) \\
    & ReAct & 131 & agent\_mismatch: 74 (31.8\%), parameter\_error: 23 (9.9\%), early\_stop: 3 (1.3\%) \\
\bottomrule
\end{tabularx}
\caption{Per-model FPoF breakdown across all five planning methods. For each model--method pair, we report the number of perfect matches (out of 233 tasks) and the three most frequent first-error types. Percentages are relative to the 233 total tasks.}
\label{tab:fpof_permodel}
\end{table*}
}

{
\setlength{\tabcolsep}{2pt}
\renewcommand{\arraystretch}{0.92}
\scriptsize
\begin{table*}[t]
\centering
\begin{tabularx}{\textwidth}{@{}>{\raggedright\arraybackslash}p{0.17\textwidth}>{\centering\arraybackslash}p{0.085\textwidth}>{\centering\arraybackslash}p{0.085\textwidth}>{\raggedright\arraybackslash}X@{}}
\toprule
\textbf{Model} & \textbf{Method} & \textbf{Perfect} & \textbf{Top-3 FPoF Error Types} \\
\midrule
Qwen3-Max   & DP & 130 & agent\_mismatch: 65 (27.9\%), parameter\_error: 24 (10.3\%), halluc.\ steps: 13 (5.6\%) \\
    & CoT & 146 & agent\_mismatch: 51 (21.9\%), parameter\_error: 27 (11.6\%), halluc.\ steps: 5 (2.1\%) \\
    & ToT & 149 & agent\_mismatch: 49 (21.0\%), parameter\_error: 28 (12.0\%), halluc.\ steps: 4 (1.7\%) \\
    & RAP & 147 & agent\_mismatch: 43 (18.5\%), parameter\_error: 28 (12.0\%), early\_stop: 7 (3.0\%) \\
    & ReAct & 145 & agent\_mismatch: 52 (22.3\%), parameter\_error: 27 (11.6\%), early\_stop: 5 (2.1\%) \\
\midrule
Qwen3-Max-Thinking   & DP & 134 & agent\_mismatch: 61 (26.2\%), parameter\_error: 25 (10.7\%), halluc.\ steps: 13 (5.6\%) \\
    & CoT & 155 & agent\_mismatch: 46 (19.7\%), parameter\_error: 21 (9.0\%), halluc.\ steps: 7 (3.0\%) \\
    & ToT & 145 & agent\_mismatch: 54 (23.2\%), parameter\_error: 23 (9.9\%), halluc.\ steps: 8 (3.4\%) \\
    & RAP & 143 & agent\_mismatch: 45 (19.3\%), parameter\_error: 28 (12.0\%), halluc.\ steps: 10 (4.3\%) \\
    & ReAct & 143 & agent\_mismatch: 55 (23.6\%), parameter\_error: 28 (12.0\%), early\_stop: 5 (2.1\%) \\
\midrule
Qwen3.5-27B   & DP & 136 & agent\_mismatch: 56 (24.0\%), parameter\_error: 28 (12.0\%), halluc.\ steps: 12 (5.2\%) \\
    & CoT & 145 & agent\_mismatch: 47 (20.2\%), parameter\_error: 32 (13.7\%), halluc.\ steps: 6 (2.6\%) \\
    & ToT & 138 & agent\_mismatch: 51 (21.9\%), parameter\_error: 25 (10.7\%), halluc.\ steps: 17 (7.3\%) \\
    & RAP & 137 & agent\_mismatch: 41 (17.6\%), parameter\_error: 28 (12.0\%), halluc.\ steps: 15 (6.4\%) \\
    & ReAct & 138 & agent\_mismatch: 49 (21.0\%), parameter\_error: 26 (11.2\%), halluc.\ steps: 19 (8.2\%) \\
\midrule
Gemma-4-31B   & DP & 155 & agent\_mismatch: 48 (20.6\%), parameter\_error: 27 (11.6\%), halluc.\ steps: 3 (1.3\%) \\
    & CoT & 147 & agent\_mismatch: 57 (24.5\%), parameter\_error: 25 (10.7\%), early\_stop: 2 (0.9\%) \\
    & ToT & 136 & agent\_mismatch: 63 (27.0\%), parameter\_error: 27 (11.6\%), halluc.\ steps: 5 (2.1\%) \\
    & RAP & 118 & parameter\_error: 73 (31.3\%), agent\_mismatch: 31 (13.3\%), halluc.\ steps: 6 (2.6\%) \\
    & ReAct & 145 & agent\_mismatch: 56 (24.0\%), parameter\_error: 25 (10.7\%), halluc.\ steps: 6 (2.6\%) \\
\bottomrule
\end{tabularx}
\caption{Per-model FPoF breakdown across all five planning methods (continued, part 2 of 4).}
\end{table*}
}

{
\setlength{\tabcolsep}{2pt}
\renewcommand{\arraystretch}{0.92}
\scriptsize
\begin{table*}[t]
\centering
\begin{tabularx}{\textwidth}{@{}>{\raggedright\arraybackslash}p{0.17\textwidth}>{\centering\arraybackslash}p{0.085\textwidth}>{\centering\arraybackslash}p{0.085\textwidth}>{\raggedright\arraybackslash}X@{}}
\toprule
\textbf{Model} & \textbf{Method} & \textbf{Perfect} & \textbf{Top-3 FPoF Error Types} \\
\midrule
Llama-3.3-70B   & DP & 90 & parameter\_error: 72 (30.9\%), agent\_mismatch: 56 (24.0\%), halluc.\ steps: 15 (6.4\%) \\
    & CoT & 137 & agent\_mismatch: 52 (22.3\%), parameter\_error: 31 (13.3\%), halluc.\ steps: 9 (3.9\%) \\
    & ToT & 97 & agent\_mismatch: 57 (24.5\%), parameter\_error: 41 (17.6\%), halluc.\ steps: 34 (14.6\%) \\
    & RAP & 67 & parameter\_error: 99 (42.5\%), agent\_mismatch: 39 (16.7\%), early\_stop: 28 (12.0\%) \\
    & ReAct & 90 & agent\_mismatch: 70 (30.0\%), parameter\_error: 49 (21.0\%), halluc.\ steps: 24 (10.3\%) \\
\midrule
Qwen3.5-9B   & DP & 107 & parameter\_error: 63 (27.0\%), agent\_mismatch: 37 (15.9\%), early\_stop: 18 (7.7\%) \\
    & CoT & 143 & parameter\_error: 49 (21.0\%), agent\_mismatch: 27 (11.6\%), early\_stop: 9 (3.9\%) \\
    & ToT & 138 & agent\_mismatch: 51 (21.9\%), parameter\_error: 24 (10.3\%), halluc.\ steps: 18 (7.7\%) \\
    & RAP & 100 & parameter\_error: 40 (17.2\%), agent\_mismatch: 39 (16.7\%), early\_stop: 39 (16.7\%) \\
    & ReAct & 153 & agent\_mismatch: 41 (17.6\%), parameter\_error: 29 (12.4\%), halluc.\ steps: 6 (2.6\%) \\
\midrule
Llama-3.1-8B   & DP & 37 & agent\_mismatch: 92 (39.5\%), parameter\_error: 87 (37.3\%), halluc.\ steps: 17 (7.3\%) \\
    & CoT & 61 & agent\_mismatch: 122 (52.4\%), parameter\_error: 29 (12.4\%), halluc.\ steps: 21 (9.0\%) \\
    & ToT & 38 & agent\_mismatch: 111 (47.6\%), parameter\_error: 43 (18.5\%), halluc.\ steps: 28 (12.0\%) \\
    & RAP & 66 & agent\_mismatch: 75 (32.2\%), parameter\_error: 57 (24.5\%), early\_stop: 15 (6.4\%) \\
    & ReAct & 62 & agent\_mismatch: 95 (40.8\%), halluc.\ steps: 37 (15.9\%), parameter\_error: 37 (15.9\%) \\
\bottomrule
\end{tabularx}
\caption{Per-model FPoF breakdown across all five planning methods (continued, part 3 of 4).}
\end{table*}
}

{
\setlength{\tabcolsep}{2pt}
\renewcommand{\arraystretch}{0.92}
\scriptsize
\begin{table*}[t]
\centering
\begin{tabularx}{\textwidth}{@{}>{\raggedright\arraybackslash}p{0.17\textwidth}>{\centering\arraybackslash}p{0.085\textwidth}>{\centering\arraybackslash}p{0.085\textwidth}>{\raggedright\arraybackslash}X@{}}
\toprule
\textbf{Model} & \textbf{Method} & \textbf{Perfect} & \textbf{Top-3 FPoF Error Types} \\
\midrule
Ministral-14B   & DP & 47 & agent\_mismatch: 67 (28.8\%), parameter\_error: 62 (26.6\%), halluc.\ steps: 57 (24.5\%) \\
    & CoT & 78 & agent\_mismatch: 61 (26.2\%), halluc.\ steps: 59 (25.3\%), parameter\_error: 34 (14.6\%) \\
    & ToT & 38 & agent\_mismatch: 82 (35.2\%), parameter\_error: 62 (26.6\%), halluc.\ steps: 51 (21.9\%) \\
    & RAP & 44 & parameter\_error: 105 (45.1\%), agent\_mismatch: 54 (23.2\%), halluc.\ steps: 25 (10.7\%) \\
    & ReAct & 118 & agent\_mismatch: 68 (29.2\%), parameter\_error: 31 (13.3\%), halluc.\ steps: 16 (6.9\%) \\
\midrule
Ministral-8B   & DP & 59 & agent\_mismatch: 82 (35.2\%), halluc.\ steps: 59 (25.3\%), parameter\_error: 33 (14.2\%) \\
    & CoT & 73 & agent\_mismatch: 92 (39.5\%), halluc.\ steps: 46 (19.7\%), parameter\_error: 21 (9.0\%) \\
    & ToT & 49 & agent\_mismatch: 77 (33.0\%), halluc.\ steps: 55 (23.6\%), parameter\_error: 51 (21.9\%) \\
    & RAP & 30 & parameter\_error: 140 (60.1\%), agent\_mismatch: 38 (16.3\%), halluc.\ steps: 20 (8.6\%) \\
    & ReAct & 103 & agent\_mismatch: 74 (31.8\%), parameter\_error: 32 (13.7\%), halluc.\ steps: 24 (10.3\%) \\
\midrule
Ministral-3B   & DP & 24 & agent\_mismatch: 136 (58.4\%), parameter\_error: 67 (28.8\%), halluc.\ steps: 6 (2.6\%) \\
    & CoT & 49 & agent\_mismatch: 140 (60.1\%), parameter\_error: 31 (13.3\%), halluc.\ steps: 13 (5.6\%) \\
    & ToT & 44 & agent\_mismatch: 82 (35.2\%), parameter\_error: 74 (31.8\%), halluc.\ steps: 17 (7.3\%) \\
    & RAP & 9 & parameter\_error: 104 (44.6\%), agent\_mismatch: 103 (44.2\%), halluc.\ steps: 10 (4.3\%) \\
    & ReAct & 86 & agent\_mismatch: 93 (39.9\%), parameter\_error: 34 (14.6\%), halluc.\ steps: 18 (7.7\%) \\
\bottomrule
\end{tabularx}
\caption{Per-model FPoF breakdown across all five planning methods (continued, part 4 of 4).}
\end{table*}
}

\section{Instruction Clash Ablation}
\label{app:clash_ablation}

To verify that the DeepSeek-V3.2 CoT collapse reflects instruction
clash rather than degraded reasoning capability, we run CoT with
API-level JSON mode enabled ($n{=}1$), which enforces structural
validity externally without modifying the reasoning prompt.
EM recovers from $36.05\%$ to $62.23\%$, surpassing DP ($57.94\%$),
while the dominant failure mode shifts from parameter\_error
($48.9\%$ of tasks) back to agent\_mismatch ($20.6\%$), closely
mirroring the DP profile ($13.7\%$ parameter error). This confirms
that the collapse is caused by competing generation objectives rather
than insufficient reasoning: when the formatting burden is
externalized, CoT's latent reasoning quality is preserved.
The no-visible-reasoning condition is instantiated by the DP baseline
by design, which outputs only the final structured plan with no
intermediate reasoning.

\section{Structural Baseline Details}
\label{app:baselines}

We include four non-LLM baselines to isolate the contribution of
language-grounded workflow generation.
\emph{Random} selects tools uniformly at random ($0\%$ EM).
\emph{Lexical-Greedy} selects tools by TF-IDF similarity to the task
description ($5.58\%$ Tool Accuracy, $0\%$ EM), confirming that
surface-level matching is insufficient for executable workflow
construction.
\emph{Shortest Path} is given the gold plan's start and end tools plus
plan length, yet achieves only $51.07\%$ Tool Accuracy and $0\%$ EM,
showing that structural hints alone cannot recover executable workflows.
\emph{Oracle-Random} receives the complete ground-truth tool sequence
but assigns parameters randomly, achieving $100\%$ Tool Accuracy yet
$0\%$ EM---isolating parameter binding as an independent execution
bottleneck even when tool selection is perfect.

\end{document}